\documentclass[journal]{IEEEtran}
\pdfoutput=1
\usepackage{cite}
\usepackage{amsmath,amssymb,amsfonts}
\usepackage{graphicx}
\usepackage{textcomp}
\usepackage{xcolor}
\def\BibTeX{{\rm B\kern-.05em{\sc i\kern-.025em b}\kern-.08em
    T\kern-.1667em\lower.7ex\hbox{E}\kern-.125emX}}
\usepackage{algorithm2e}
\makeatletter
\renewcommand{\@algocf@capt@plain}{above}
\makeatother
\usepackage{caption}
\usepackage{multicol,lipsum}
\usepackage{cuted}
\usepackage{mathtools}
\usepackage{array}
\usepackage[T1]{fontenc}
\usepackage{multirow}
\usepackage{longtable}
\usepackage{booktabs}
\usepackage[utf8]{inputenc}
\usepackage{titlesec}
\usepackage{epstopdf}
\usepackage{tikz}
\usepackage{textcomp}
\usepackage{hyperref}
\usepackage{lipsum}
\usepackage{caption,subcaption}

\usepackage{verbatim}
\usepackage{color,soul}

\newcommand\copyrighttext{%
  \footnotesize \textcopyright 2023 IEEE.  Personal use of this material is permitted.  Permission from IEEE must be obtained for all other uses, in any current or future media, including reprinting/republishing this material for advertising or promotional purposes, creating new collective works, for resale or redistribution to servers or lists, or reuse of any copyrighted component of this work in other works.
  DOI: \href{https://doi.org/10.1109/TEVC.2022.3178299}{<10.1109/TEVC.2022.3178299>}}
\newcommand\copyrightnotice{%
\begin{tikzpicture}[remember picture,overlay]
\node[anchor=south,yshift=10pt] at (current page.south) {\fbox{\parbox{\dimexpr\textwidth-\fboxsep-\fboxrule\relax}{\copyrighttext}}};
\end{tikzpicture}%
}

\begin{document}

\title{Evolutionary Computation in Action: \\Feature Selection for  Deep Embedding Spaces \\of Gigapixel Pathology Images}

\author{Azam~Asilian~Bidgoli,
        Shahryar~Rahnamayan,
        Taher~Dehkharghanian,
        Abtin~Riasatian,
        and~H.R.~Tizhoosh
\thanks{A. Asilian Bidgoli, S. Rahnamayan, and T. Dehkharghanian are with Nature Inspired Computational Intelligence (NICI) Lab,  Ontario Tech University, Oshawa, Canada, Corresponding author's e-mail: azam.asilianbidgoli@uoit.ca.}
\thanks{A. Riasatian is with Laboratory for Knowledge inference in Medical Image Analysis (Kimia Lab), University of Waterloo,  Ontario,  Canada.}
\thanks{H.R. Tizhoosh is with Laboratory for Knowledge inference in Medical Image Analysis (Kimia Lab), University of Waterloo,  Ontario,  Canada and 	Mayo Clinic, Rochester, MN, United States.
}
}
\markboth{Journal of \LaTeX\ Class Files,~Vol.~14, No.~8, August~2015}%
{Shell \MakeLowercase{\textit{et al.}}: Bare Demo of IEEEtran.cls for IEEE Journals}

\maketitle
\copyrightnotice

\begin{abstract}

One of the main obstacles of adopting digital pathology is the challenge of efficient processing of hyperdimensional digitized biopsy samples, called \emph{whole slide images} (WSIs). Exploiting deep learning and introducing compact WSI representations are urgently needed to accelerate image analysis and facilitate the visualization and interpretability of pathology results in a postpandemic world. In this paper, we introduce a new evolutionary approach for WSI representation based on large-scale multi-objective optimization (LSMOP) of deep embeddings. We start with patch-based sampling to feed \emph{KimiaNet} , a histopathology-specialized deep network, and to extract a multitude of feature vectors. Coarse multi-objective feature selection uses the reduced search space strategy guided by the classification accuracy and the number of features. In the second stage, the frequent features histogram (FFH), a novel WSI representation, is constructed by multiple runs of coarse LSMOP. Fine evolutionary feature selection is then applied to find a compact (short-length) feature vector based on the FFH and contributes to a more robust deep-learning approach to digital pathology supported by the stochastic power of evolutionary algorithms.
We validate the proposed schemes using The Cancer Genome Atlas (TCGA) images in terms of WSI representation, classification accuracy, and feature quality. Furthermore, a novel decision space for multicriteria decision making in the LSMOP field is introduced. Finally, a patch-level visualization approach is proposed to increase the interpretability of deep features. The proposed evolutionary algorithm finds a very compact feature vector to represent a WSI (almost 14,000 times smaller than the original feature vectors) with 8\% higher accuracy compared to the codes provided by the state-of-the-art methods.

\end{abstract}


\IEEEpeerreviewmaketitle

\section{Introduction}

\IEEEPARstart{H}{istopathology}     is the study of changes in tissue samples under a light microscope at different magnification levels. Because of the invasive nature of biopsy procedures, histopathology inspection is usually needed when an ultimate diagnosis is needed. Histopathologic examination is the gold standard method for diagnosing many diseases, e.g., cancer diagnosis. Furthermore, pathology has a broader application in diagnosing many metabolic and autoimmune disorders, inflammations, and infections \cite{Doley, brunt_2010, emre_bilge, goodman_ishak}. Digitization of conventional glass slides into whole slide images (WSIs) has opened a new branch for the implementation of computer vision techniques utilized in computer-aided diagnosis (CAD) systems \cite{Pantanowitz2015}.
WSIs are generally gigapixel images (e.g., 100k$\times$100k pixels), from which smaller size sample images should be taken out to represent each WSI \cite{KALRA_yottixel}. Hence, a mosaic, i.e., a set of patches, should be extracted from WSI regions to characterize each sample. However, a large number of patches makes it difficult to interpret any downstream task in computational pathology.  \\
In recent years, deep neural networks (DNNs) have become the method of choice for feature extraction, which is in an integral part of digital pathology~\cite{Saritha2019}. The DNN-derived feature vectors used for image representation usually have a length of 512 to 2048. Consequently, storing a large number of DNN-derived feature vectors has a substantial memory burden; moreover, DNNs are  slow processing algorithms, which often make them impractical in real-world applications. Hence, decreasing the length of a DNN-derived feature vector would make it plausible to increase the number of DNN-derived feature vectors per WSI, which would eventually improve WSI representation. Additionally, it has been shown that it is possible to attribute humanly perceivable visual patterns to specific DNN-derived features. A smaller DNN-derived feature vector would allow for the investigation of this property to make WSI representation more interpretable for expert users. In light of this motivation, feature selection is a well-known technique that can be employed to reduce the size of a DNN-derived feature vector accordingly.
Since DNNs commonly generate a large set of features, especially for extremely large images such as WSIs, the feature selection task on large datasets can be modeled as an expensive large-scale multi-objective  optimization (LSMOP) problem. On the other hand, the major objective of the modeled optimization task can be maximizing the accuracy of image retrieval, whereas the number of features can be reduced as much as possible. Although feature selection can enhance the accuracy of the classification task
and decrease the computational complexity, an extreme reduction in relevant
features will degrade the accuracy~\cite{xue2012particle}. Accordingly, feature selection can be modeled as a multi-objective large-scale optimization problem with two objectives, namely, maximization of the accuracy of image retrieval and minimization of the number of selected features.
In one of the recently published state-of-the-art studies, the authors trained a DNN, called KimiaNet~\cite{riasatian2021fine}, to extract the features from WSIs of TCGA~\cite{gutman2013cancer}. In this study, an evolutionary multi-objective feature selection (EMOFS) algorithm is proposed to find an optimal set of DNN-derived features for each category of tumor type. In fact, for this goal, the features are transferred from a DNN to an evolutionary algorithm to optimize the output results of the DNN. The optimization of a DNN is a crucial, challenging and sophisticated due to the computational complexity; however, optimization of the DNN output can alleviate the suboptimality of DNNs.

Evolutionary algorithms have been widely employed as feature selectors in many machine learning- or data mining-related applications~\cite{xue2015survey,hosseini2019evolutionary}.
The stochasticity in such algorithms is generally a challenging issue that may affect their utilization in some research communities or practical environments. Despite the power of these algorithms in exploration and exploitation of the search space to find the optimal/desired solution, especially in large-scale problems~\cite{deb2016breaking}, the role of randomness in different steps of the process possibly leads to a different solution being generated after each run of the algorithm.
This specific nature of evolutionary algorithms employed as feature selectors has a negative impact on the stability of the selected subset.
 The stability of the feature selection algorithm provides confidence in the robustness of the features to data and function perturbations (i.e., algorithm settings)~\cite{pes2017exploiting}. Therefore, it is valuable to propose a method to alleviate the unsteadiness and to select a more reliable and significant feature set that is robust to selection bias~\cite{ambroise2002selection}.
    Two fundamental factors that increase the instability in feature selection algorithms are the large-scale search space and stochasticity, which are the main characteristics of the DNN-derived feature selection approach utilized in this study; these characteristics are not only properly addressed but also interestingly inversed into  an enhancement factor. To satisfy the stability condition, an approach is required to reduce the impact of these factors on the finalized feature subset.

In this study, we rely on the benefits of the stochasticity of evolutionary algorithms to produce a more robust feature subset. Multiple independent runs of the multi-objective algorithm generate a set of Pareto front solutions from which useful knowledge can be extracted to establish a stable feature subset. The process of knowledge discovery from the Pareto front of a multio-bjective optimization problem is called innovization~\cite{deb2014integrated}. Deb and Srinivasan~\cite{deb2006innovization} introduce it as ``a new design methodology in the context of finding new and innovative design principles by means of optimization techniques''. In multi-objective optimization, innovization is commonly employed to discover knowledge from the Pareto front solutions to decipher design rules that are meaningful to a human designer. Knowledge transfer from a solved problem can guide an optimizer to achieve higher efficiency and performance for solving similar problems.
   In this study, the search space is reduced by selecting a number of prominent features using the useful knowledge transferred across the first level of the optimization process on the original search space.\\
   The frequency of feature occurrence in the different subsets of the Pareto front and those that appear in multiple independent runs of the algorithm can be interpreted as a Darwinian-based indicators of feature prominence. A frequent features histogram (FFH) is constructed to generate the new search space for the second level of optimization, namely, fine optimization (i.e., two-stage evolutionary feature selection). On the other hand, the search space of the second-level optimization process is decreased significantly to solve the instability~\cite{yang2013stability} of feature selection and to augment the accuracy of the image search.

The major contributions of this study include the following: 1) to the best of our knowledge, this is the first time that WSIs are presented by an extremely compact code (i.e., 13,800 times shorter than the initial length) compared to the state-of-the-art algorithms. 2) To find an optimal feature vector, a strategy for decreasing the search space is proposed by defining a constraint on objective values, which eliminates some parts of the search space. Accordingly, the evolutionary operators are modified for the multi-objective optimization algorithm to satisfy the corresponding constraint. 3) A two-level feature selection framework is proposed, and it injects the knowledge discovered from the first level to the second level not only to decrease the dimensionality of the optimization problem but also to yield a robust feature vector. 4) By providing a compact WSI code, we visualize the image search results with a new method for region matching, which is crucial for WSI visualization in digital pathology.
Consequently, this study provides a general framework that utilizes an evolutionary component for any pretrained DNN to tackle its drawbacks, such as the suboptimal design and/or specific lack of sufficient training data in digital pathology, and to train an end-to-end deep network to extract organ-based feature vectors.

The remaining sections of this paper are organized as follows: A background review is provided in Section II. Section III presents a detailed explanation of the proposed multi-objective feature selection approach. Using the TCGA dataset, the performance of the proposed method is investigated in Section IV. Finally, the paper is concluded in Section VI.

\section {Background Review}

The gigapixel size of histopathology images motivates researchers to study novel approaches for decreasing the size of representative vectors; otherwise, their processing would be very complicated or even impossible in practical settings .
It is extremely crucial to address this issue because the common way of representing a pathology image is to extract a large number of patches from a WSI, each of which requires a feature vector as its representative vector.
 In recent years, pretrained DNNs have become the method of choice for feature extraction of histopathology images utilized in digital pathology, including content-based image retrieval (CBIR)~\cite{kalra2020pan, hegde_2019}.
Luigi, Yottixel, and SMILY are three examples of CBIR systems that use pretrained DNNs for DNN-derived feature extraction. Luigi is an application that uses deep texture representations created by a pretrained DNN and exploits the nearest neighbor search to retrieve similar cancer histopathology images \cite{luigi}. Yottixel represents each WSI with a mosaic set of tissue patches. Then, DNN-derived features extracted from each tissue patch are converted into binary barcodes, which require less storage space \cite{KALRA_yottixel}. SMILY applies a pretrained DNN to tissue images at 10x magnification to build a search dataset using an embedding computational module \cite{hegde_2019}. All of the abovementioned CBIR systems have utilized WSIs taken from The Cancer Genome Atlas (TCGA) repository\cite{tcga_2015}.

Despite the successful employment of DNNs, DNN-derived features must be efficiently optimized in many applications. Deep selection has been conducted in some recent successful studies. In~\cite{bar2018chest}, the authors applied Kruskal–Wallis feature selection on a set of DNN-derived features along with a set of traditional handcrafted features to select the best combination. The features were extracted from chest pathology images. The features were reduced using the minimum redundancy maximum relevance algorithm on the output of three deep models to select the best feature set extracted from X-ray images~\cite{tougaccar2020deep}. Accordingly, the authors combined the selected features from independent deep models to provide an efficient feature set. In a similar study, Ozyurt developed a feature selection framework on DNN-derived features generated by several well-known pretrained networks~\cite{ozyurt2019efficient}. The features are selected based on the ReliefF algorithm. In~\cite{mirzaei2020deep}, the DNN-derived features extracted by a teacher autoencoder are fed into a shallow student network to rank the low-dimensional DNN-derived features. In the last step of the proposed framework, the top ranked features are selected by the weights inspired by the student network. A comprehensive comparative study was conducted by evaluating eleven feature selection algorithms on three conventional DNN algorithms, i.e., convolutional neural networks (CNNs), deep belief networks (DBNs), recurrent neural networks (RNNs), and three recent DNNs, i.e., MobilenetV2, ShufflenetV2, and Squeezenet~\cite{chen2020feature}. The experimental data supported their hypothesis that feature selection algorithms may improve DNN performance.

One of the major issues that should be addressed in a feature selection task is the instability of the feature selection results.
The stability of feature selection has been studied in two major aspects, including 1) proposing various approaches for improving the stability of feature selection algorithms and 2) defining various metrics for measuring the stability of feature selection~\cite{yang2013stability}.
There are various methods to solve the instability of the feature selection algorithm, which are discussed in~\cite{yang2013stability,khaire2019stability}. The ensemble feature selection technique is based on the idea of \textit{collective intelligence}, which demonstrates that a population is collectively
more intelligent than each individual for problem solving and decision making~\cite{bidgoli2020collective}. In general, ensemble methods work based on creating a set of different selectors (ensemble components) and aggregating the results of the different selectors into a single final decision (i.e., result-level fusion).
One of the popular frameworks for ensemble methods is to combine multiple selectors that apply the same algorithm on different perturbed versions of the original data~\cite{pes2017exploiting}. In another category of ensemble-based feature selection methods, different algorithms or one algorithm with different control parameters are exploited on the same data to collaboratively select a set of aggregated solutions for one robust subset of features~\cite{chiew2019new}.
Accordingly, the different runs of the algorithm with the optimizer can provide useful knowledge for finding more robust solutions.

The knowledge transfer for feature selection is valuable from the following several key perspectives: 1) The feature information strategy: this approach is one of the metrics to measure the prominence of each feature based on some assessment standards~\cite{khaire2019stability}. Then, stable features can be selected from these highly outstanding features. For this purpose, the frequency can be an indicator to assess the significance of a feature. Therefore, the problem is converted to a low-scale optimization problem. Moreover, since features in low-dimensional space are less sensitive to small data perturbations, a subset of frequent features can be more stable than a set selected from all features~\cite{khaire2019stability}. 2) An Ensemble of the feature selectors: an ensemble method combines the results generated by the same feature selection method.

\begin{figure*}[!ht]
\centering
\includegraphics[width=0.8\textwidth]{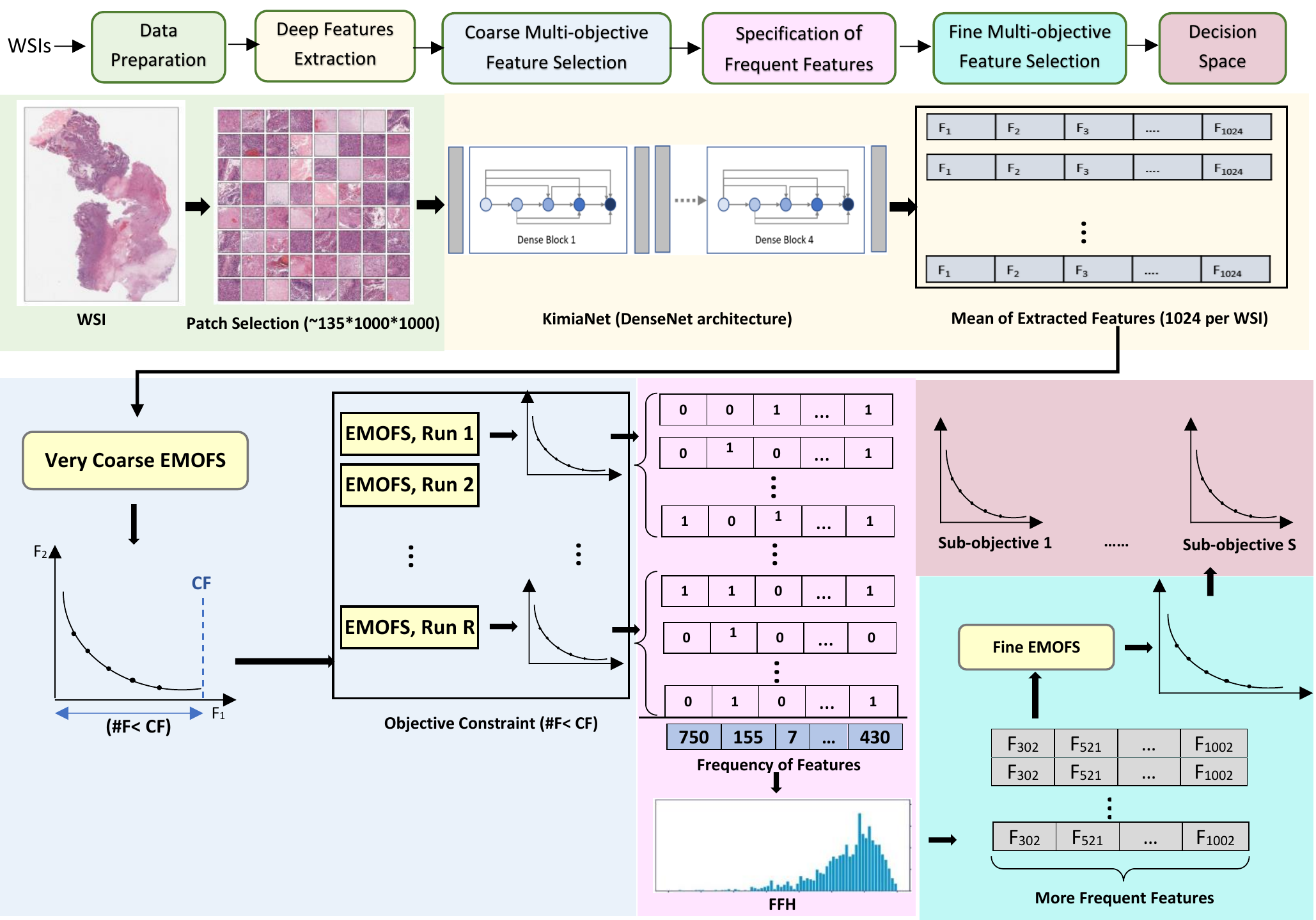}

\caption{The overall structure of the  proposed framework.  Each color illustrates a specific section of the proposed framework. The process of EMOFS is represented as a separate block.  }
\label{Fig:Process}
\end{figure*}

\section{Proposed Two-Stage Multi-Objective Feature Selection on Histopathology Images}
The proposed method consists of several major components, as illustrated in the high-level diagram in Fig.~\ref{Fig:Process}. Before the detailed explanation of each component, an overall overview is provided as follows:
\begin{enumerate}
    \item Data Preparation: Each WSI is divided into nonoverlapping patches to select more cellular regions.
    \item Deep Feature Extraction: From each patch, the representative features are extracted by a DNN trained on histopathology images. Thereafter, the mean of the feature vectors is calculated for each patch, resulting in one feature vector of size 1024 for the entire WSI.
    \item Coarse multi-objective Feature Selection: The best subset of features is selected based on two conflicting objectives: maximization of classification accuracy and minimization of the number of mean features.
    \item Specification of Frequent Features: The most frequent features from 10 runs of the coarse optimization phase are selected to construct the shrunken search space for the second stage of optimization.
    \item Fine Multi-objective Feature Selection: Multi-objective feature selection is applied one more time on the most frequent features to pick the optimal and stable feature subset.
    \item Decision Space: To facilitate decision making after multi-objective  optimization, a new decision space is constructed by decomposing the objectives into subobjectives.
\end{enumerate}

\textbf{Data Preparation and Feature Extraction}. As indicated, the process begins with the data preparation step explained in detail in~\cite{riasatian2021fine}. The authors extracted image patches of size $1000 \times 1000$ pixels at $20\times$ magnification with no overlap from each WSI. This method can be seen as a way of ``summarizing a WSI'' by using a small subset of representative patches. In the next step, as presented in Fig.~\ref{Fig:Process}, the representative features should be extracted from each patch. Recently, KimiaNet~\cite{riasatian2021fine} was trained to extract features from patches to distinguish the 30-class tumor primary diagnosis defined in the repository of TCGA. KimiaNet is a network with a DenseNet topology that consists of four dense blocks with preceding convolutional and pooling layers. The network is fully trained using 242,202 and 24,646 patches in the training and validation datasets \cite{riasatian2021fine}. The extracted features are evaluated for image retrieval from the repository of TCGA.
A feature vector of size 1024 represents each patch. On average, a WSI is represented by 135 patches. Therefore, for a WSI, 138,240 ($=135\times1024$) values are needed to describe its visual content. We realized that based on center-based sampling, using the mean of the feature vectors (MFV) over all patches of a WSI decreases the amount of representative data (i.e., 135 times smaller), while the accuracy of image retrieval increases. Center-based sampling is a concept introduced first in optimization, which indicates that the points that are closer to
the center of the search space have a higher chance of being closer to an unknown solution in a black-box problem~\cite{rahnamayan2009center,hiba2019cgde3}. In our application, the center of the features for patches is a proper representative of all patch es. This not only decreases the vector size for WSI representation but also enhances the image retrieval accuracy.
In this way, each WSI is represented by only one vector of size 1,024 instead of 135 vectors. However, the final goal is to select the best subset of features for more efficient processing in digital pathology. Hence, two levels of optimization, namely, coarse and fine optimizations, are conducted on the features of TCGA images to construct a Pareto front of robust subsets of features.
The repository of TCGA includes 12 tumor type categories, each of which encompasses several primary diagnoses (tumor subtype), resulting in 30 primary diagnoses overall.
The main task for image retrieval would be a ``vertical search'' in which the most similar samples are found from a specific tumor type and not from the entire dataset.
Motivated by this, for each tumor type, an optimal set of highly relevant features can be selected among the 1024 features to accurately classify the primary diagnosis related to the corresponding tumor type. However, the DNN is trained on all types of primary diagnoses. Accordingly, by an evolutionary algorithm, the knowledge inherited from the DNN is transferred to a new task to discriminate only a subset of tumor subtypes. Initial experiments showed that there is a large number of irrelevant features among the extracted features when used for search and classification. This is due to two potential reasons. First, since the network is trained universally to detect all types of tumors (i.e., 30 classes), a very small subset of features can be selected for a particular tumor category, including a maximum of 4 primary diagnoses. Second, the trained DNNs are possibly not optimal in terms of structure and topology; that is, they may extract a large amount of irrelevant information during the training process. Thus, the proposed method remedies the suboptimality of extracted knowledge by selecting a subset of evolutionary features for each tumor category. Downstream tasks, such as retrieval and classification, can then be accomplished by this subset more effectively and efficiently. Additionally, it is worth mentioning that the lack of sufficient data in each category is a crucial obstacle to training an end-to-end deep model to extract a compact code for each WSI to distinguish the cancer types of a tumor site.

In light of the aforementioned points, feature selection can be conducted on each site category (instead of the entire dataset) to find a specialized optimal subset of features. The evolutionary feature vectors (EFVs) from coarse optimization increase the efficiency of image retrieval by choosing the F1-score as the first objective of optimization. The F1-score is a harmonic mean of precision and recall measures and is defined as follows:

\begin{equation}
\label{F1}
\mathit{F1\mbox{-}score}= 2\times \frac{Precision\times Recall}{Precision+ Recall},
\end{equation}

\begin{equation}
\label{PR}
\mathit{Precision}= \frac{True Positive}{True Positive+False Positive}
\end{equation}

\begin{equation}
\label{RE}
\mathit{Recall}= \frac{True Positive}{True Positive+False Negative}
\end{equation}

On the other hand, the number of features can be considered the second objective. 
The curse of dimensionality is a critical issue when constructing a classification model on high-dimensional data~\cite{debie2019implications}. A shorter code can accelerate the retrieval process among numerous gigapixel images and decrease the memory requirements.
 Although fewer features decrease the computational cost, insufficient
relevant features will prevent the learning algorithm from classifying WSIs effectively~\cite{bidgoli2021reference}. Accordingly, the size of an EFV and the classification accuracy are considered two conflicting objectives. In the following subsections, the details of both levels of the proposed multi-objective  feature selection method are explained.

\subsection{Coarse Multi-Objective Feature Selection}

The optimization problem is formulated as a large-scale and expensive problem because the dimension of the search space is 1024 and each image must be compared with a very large image set. Accordingly, a multi-objective  evolutionary feature selection method (i.e., the third block of Fig.~\ref{Fig:Process}) is proposed to select the most relevant subset among 1024 DNN-derived features. Two considered objectives are the minimum number of features and maximum image search accuracy defined by the average F1-score over classes.

\begin{figure*}[!ht]
\centering
\includegraphics[width=0.8\textwidth]{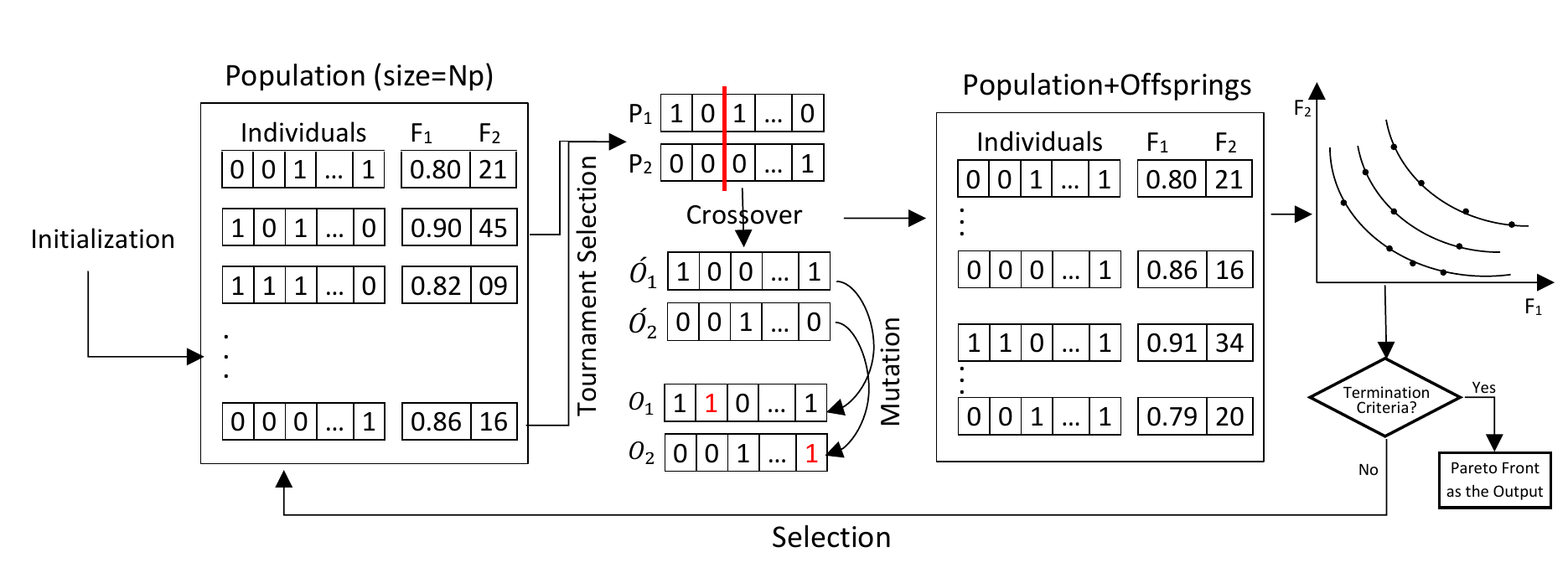}

\caption{ The process of EMOFS. Each individual is presented by a binary vector which indicates the selection (1) or non-selection (0) of a feature.  }
\label{Fig:EMOFS}
\end{figure*}

The method is inspired by the NSGA-III algorithm~\cite{deb2013evolutionary}, which is fundamentally modified to improve the performance of the algorithm for targeted feature selection. The modifications include defining a new constraint on the objective function, tailoring the generative operators to satisfy the constraint, and a new initialization scheme, all of which are explained in the following.
Generally, feature selection is inherently a binary optimization problem; thus, the genotype representation of individuals in the population are binary vectors of size 1024. Each bit indicates the status of a feature, where 0 reflects the absence of a feature and 1 is indicative of the feature's presence. The phenotype is the number of genes with a value of 1, which reflects the selected features that contribute to the next processing step. For each category of tumor sites, we accomplish optimization to select the smallest subset of features with the highest image search accuracy. After only one run of the algorithm (indicated as very coarse optimization in Fig.~\ref{Fig:Process}), we noticed that the number of selected features on the resulting Pareto front solutions does not exceed a limited value (i.e., $CF\!=\!50$). As mentioned previously, to classify at most 4 tumor subtypes out of 30, a very small set of features is needed.
This observation motivated us to run the multi-objective  evolutionary algorithm on a shrunken search space to find the subsets with less than 50 features; in fact, in this way, we reduce the cost of the optimization process. To this end, during optimization, we set a box constraint ($\#F\!<\!CF$) on the second objective, which is the number of features. As a result, the optimizer deeply explores the space within a specific interval of the corresponding objective. Obviously, the size of the search space with 1024 features is $\sum_{i=1}^{1024}C(1024,i)=2^{1024}$, where $C(1024,r)$ indicates the number of ways that the selection of $r$ features from 1024 features can be performed. This number is much larger than all possible combinations of the selection of a maximum of 50 features from the 1024 features given by $\sum_{i=1}^{50}C(1024,i)$.
Based on this consideration, we decreased the number of fitness calls and the population size to half of their original values. In this way, it is possible to reduce the search budget for the optimizer to find an optimal Pareto set. In the following, the steps of the proposed EMOFS approach are explained. Additionally, the process of EMOFS is illustrated in Fig.~\ref{Fig:EMOFS}.


\textbf{\emph{Initialization}}. The population is initialized with a set of random binary individuals. To distribute the population adequately in the objective space, we generate $N$ uniform random numbers, $NF_1, NF_2, NF_3,\dots,NF_N$ in $(0,CF)$, where $CF$ reflects the maximum number of selected features (i.e., 50 in our application). $NF_i$ indicates the number of features for the $i$th individual in the initial population. Therefore, for individual $i$, $NF_i$ bits should be selected uniformly randomly and set to 1, and certainly, the values of the other genes remain 0.
The algorithm starts with the evaluation of the initial population based on both objectives.

\textbf{\emph{Generative operators}}. The new individuals are generated by a one-point crossover and bitwise mutation. To satisfy the constraint of the second objective (i.e., $\#F\!<\!CF$), both operators are required to be tailored to reduce the number of selected features down to a feasible value for new individuals. To this end, if the number of features in the mutant individual is more than $CF$, the optimizer chooses a random set of selected features to flip their corresponding variables from 1 to 0. Suppose that the number of selected features in the mutant vector is $EF$, then at least $EF\!-CF$ features should be removed to bring the objective value into the feasible interval (i.e., objective-constraint satisfaction). Hence, a random number, $RF$ in $[EF-CF, EF]$, is selected to indicate the number of extra features that should be flipped.
Therefore, we control the mutation and crossover to avoid violating the defined objective constraint.

\textbf{\emph{Selection.}} Based on the next common step of the evolutionary multi-objective  algorithms, the best candidate solutions among the previous and current population should be selected for the next generation.
The NSGA-III selection strategy is employed to choose the top ranked feature subsets based on the nondominated sorting (NDS) algorithm, and the reference lines are distributed uniformly in the search space to provide well-spread out candidate solutions.
The pseudocode of the algorithm is presented in Algorithm~\ref{alg1}.

\begin{algorithm}
\SetAlgoLined
\SetKwInOut{Input}{inputs}\SetKwInOut{Output}{output}
 \Input{ $D$: Number of features, $CF$: Box constraint on number of selected features, $EVmax$: Maximum number of fitness function evaluations, $NP$: Population size}
 \Output{ $NS$: Non-dominated solutions}
 \BlankLine
\tcp{Initialization}
$NFs$= $NP$ uniform random numbers in $[1,CF]$

\For{$i\leftarrow 1$ \KwTo $NP$}
 {
 $POP(i)$= A uniform binary vector with $NFs(i)$ bits of 1\;
  }

$Eval=1$\;
Evaluate each individual in terms of objectives which are  number of features and F1-score\;
\tcp{Main algorithm}
\While{$Eval<EVmax$}{
$NPOP=POP$\;
Generate $NP$ individuals by binary Genetic operators and add them to $NPOP$ based on crossover and mutation operators\;
\tcp{Fitness Evaluation}
Evaluate each individual in terms of objectives which are  number of features and F1-score\;
 \tcp{Selection strategy}
 $POP$ = Select $NP$ individuals from  $NPOP$ based on NSGA-III algorithm\;
 $Eval=Eval+1$\;
}
Determine Non-dominated solutions as $NS$\;
\caption{Pseudo-code for the multi-objective feature selection inspired from NSGA-III algorithm.}\label{alg1}
\end{algorithm}

\subsection{Fine Multi-Objective Feature Selection based on the Frequency of Features}

Although the stochastic nature of the evolutionary algorithms can be interpreted as a weakness, the useful knowledge extracted from multiple independent runs may lead to finding more robust and better solutions. In other words, innovization is performed by investigating Pareto fronts resulting from evolutionary feature selection and more specifically on the frequency of feature subsets. The optimizer exploits the discovered empirical knowledge to find high-quality solutions in terms of accuracy and stability.

Before we explain the method, an explicit definition of the frequency must be given.
As mentioned previously, we obtained the EFVs using $R$ independent runs of multi-objective  optimization. As illustrated in the fourth block of Fig~\ref{Fig:Process}, a set of candidate solutions (i.e., a subset of features) emerges from each run. Among the resultant subsets, there are some common features. The $i$th cell of the blue vector indicates the number of resultant subsets (i.e., frequency) containing feature $i$. The investigation of the frequency of features demonstrates some inspiring patterns on the Pareto fronts. Two types of frequencies are observed: the duplication of features in different subsets on a Pareto front resulting from one run and the repetition of selected features on the results of different runs (i.e., different Pareto fronts). The aim of using the frequency is to achieve a robust feature subset that alleviates the stochasticity of the approach. However, the diversity of efficient feature subsets reveals the high multimodality of the optimization problem, which possesses many optimal/suboptimal candidate solutions.\\
Despite the key role of randomness in evolutionary algorithms, which are included in initialization and generative operators, the selection of some features repeatedly indicates their obvious prominence. The frequency of a reselected feature indicates its impact on increasing the accuracy of the image search as one of the objectives.
To calculate the prominence of each feature, the scores assigned based on the frequency generate the FFH, in which each bin represents the score of each feature. Since the occurrence of a feature in different runs is more valuable than that in different subsets of the Pareto front, we allocate more weight to the frequency in each run. For this purpose, we calculate the score ($FreqScore$) of a feature in terms of the frequency as follows:

 \begin{equation}
\label{FR}
FreqScore(f)=\sum_{r=1}^{R}\delta\left(1+\frac{Fr(r,f)}{R}\right),
\end{equation} 
where $R$ is the number of runs of the algorithm and $Fr(r,f)$ is the number of subsets on $r$-th Pareto front in which  feature $f$ exists. Additionally, $\delta$ is an indicator for the Pareto fronts which consists of feature $f$.
\begin{equation*}
\delta=
\begin{cases}
1  &  \text {if} \hspace{0.3cm}  Fr(r,f)>0\\
0   &  \text {if} \hspace{0.3cm} Fr(r,f)=0
\end{cases}\hspace{2cm}
\end{equation*}
  The occurrence of a feature on a run earns one score whereas  the number of occurrences in the different subsets on the same Pareto front is divided by the number of runs to allocate a lower score (i.e., $1/R$).

The investigation of the quality of frequent features reveals that searching inside the promising region generated by those features can result in solutions that are more efficient than the EFVs acquired by the coarse optimization process. The idea of creating a shrunk search space for a large-scale optimization problem is presented; for this problem, we use some innovization-based knowledge from the coarse optimization process to determine the more promising area of the search space and correspondingly compute the frequency-based feature vectors (FFVs).
Selection using FFH can be an efficient approach to provide high-quality subsets in terms of image search accuracy and reliability. Therefore, fine-level optimization can be conducted on a set of frequent features that have higher scores computed by Eq.~\ref{FR}. For this purpose, as presented in the fifth block of Fig~\ref{Fig:Process}, the dataset is reconstructed based on the most frequent features, and the proposed EMOFS model is rerun on this subset of features.

The applied the multi-objective  optimization algorithm for fine optimization is similar to what we proposed for coarse optimization. Apart from the dimension of the problem, which is reduced to the number of top frequent features ($NFF$), the other parts of the formulation of the optimization problem are similar to those of coarse optimization. In the second level of optimization, the objective constraint (i.e., $\#F\!<C\!F$) is ignored as the dimension of the problem is sufficiently low. The pseudocode of the fine optimization algorithm is provided in Algorithm~\ref{alg2}. In this phase, the optimizer tries to select the best subsets among $NFF$ frequent features according to both objectives: the minimum number of features and the maximum average F1-score. As mentioned before, the result of this level of optimization is expected to be more accurate and more robust, as presented in the experiments.


\begin{algorithm}
\SetAlgoLined
\SetKwInOut{Input}{inputs}\SetKwInOut{Output}{output}
 \Input{ $Dataset$: Deep features, $D$: Number of features, $CF$: Constraint on number of selected features, $NFF$: number of top frequent features, $EVmax_C$: Maximum number of function evaluations for coarse optimization, $EVmax_F$: Maximum number of function evaluations for fine optimization, $NP$: Population size, $R$: Number of runs}
 \Output{ $NS$: Non-dominated solutions}
 \BlankLine
 \tcp{Generating frequent features}
 \For{$r\leftarrow 1$ \KwTo $R$}
 {
 $Pareto front(r)$=  Algorithm 1 ($Dataset$, $D$, $CF$, $EVmax_C$,$NP$)\;
 \For{$f\leftarrow 1$ \KwTo $D$}
 {
 \If{$f \in PF(r)$}{
 $FreqScore(f)=\sum_{r=1}^{R}\delta\left(1+\frac{Fr(r,f)}{R}\right)$\;
 }
 }
 }
 $FreqScore$=Sort($FreqScore$)\;
 $Dataset$=$Dataset$($Freq$(1:$NFF$))\; 
  $NS$=Algorithm 1 ($Dataset$, $NFF$, $NFF$, $EVmax_F$,$NP$)\;
\caption{Pseudo-code for the fine multi-objective optimization conducted on the most frequent features Again inspired from NSGA-III. }\label{alg2}
\end{algorithm}

\section{Experimental Results}
In this section, we explain the conducted experiments to assess the proposed two-stage multi-objective  feature selection method for gigapixel pathology images.

In this section, we explain the conducted experiments to assess the proposed two-stage multi-objective  feature selection method for gigapixel pathology images.

\subsection{Dataset and Implementation Details}
We used the repository of TCGA~\cite{gutman2013cancer}, which is the largest publicly available histopathology dataset.
 Since the size of histopathology images is very large (e.g., 100k$\times$100k pixels), a common method to facilitate the processing of such images is to split the WSI into smaller subimages called \emph{patches}. On average, a WSI may be represented by 135 patches every 1000$\times$1000 pixels if proper patching and clustering techniques are used \cite{KALRA_yottixel,kalra2020pan}. \\
For each patch, a feature vector of size 1024 is extracted by KimiaNet trained using soft labels on high cellularity patches~\cite{riasatian2021fine}. During the preprocessing steps, the authors divided the patches into 7126, 741 and 744 diagnostic WSIs for the training, validation, and test datasets, respectively. The network is trained on a set of patches to extract the features based on 30 primary diagnoses. As a result, a WSI is represented by 135 (on average) feature vectors of size 1024. A WSI is represented by an MFV with only 1024 values. An MFV is meaningful in our application because KimiaNet has been trained on \emph{high cellularity} patches, which enforces homogeneity. \\
As previously mentioned, the feature vectors are extracted to discriminate the primary diagnosis over the entire dataset (i.e., the discrimination of 30 cancer types). However, the main aim of an image retrieval system is to retrieve the most similar images from a specific category of tumor type.
A tumor site categorization is established on the TCGA dataset \cite{cooper2018pancancer,kalra2020pan}, which results in 12 tumor sites, including the brain, breast, endocrine, gastrointestinal tract, gynecological tissue, hematopoietic tissue, melanocytes, mesenchymal, liver, prostate/testis, pulmonary tissue, and urinary tract. All tumor sites except the breast, hematopoietic tissue, and mesenchymal sites consist of more than one subtype on which the optimization is applied. Table~\ref{Data} represents the defined ID for each primary diagnosis, the related number of patients, and the details of tumor categorization. \\
In this study, 12 independent tumor-based optimization problems are defined to decrease the number of features and to select the optimal subset specialized for each tumor type category. The $CF$, $NFF$, and population size parameters are set to 50, 30, and 50, respectively. The number of fitness evaluations for coarse and fine optimization are set to 512,000 and 1000$\times$ 30 =30,000, respectively. To evaluate the candidate solutions during the optimization process, the validation set images are searched among the training set using the $k$-nearest neighbor approach~\cite{pandey2017comparative}. The three most similar images based on the Euclidean distance are retrieved to calculate the average F1-score over all primary diagnoses as the first objective. The number of selected features is the second objective, which is rescaled to $[0,1]$.
The experiments are performed with 31 independent runs according to statistical studies, and 30 is considered the minimum sample size in some forms of statistical analysis with the required margin of error and confidence level~\cite{islam2018sample,johanson2010initial}. Accordingly, most of the evolutionary computation studies use this number of runs for their experiments. Therefore, the 31 independent runs of the algorithm yield 31 Pareto fronts for assessment.
The final feature subsets on the Pareto fronts are evaluated on the test set. Finally, the Wilcoxon statistical test~\cite{demvsar2006statistical} is conducted to investigate the significance of the acquired results. For this purpose, the Wilcoxon test is applied to the results of the 31 runs for each tumor type separately. In this way, the statistical test evaluates the significance of the winning method on each tumor type by considering the variance on the 31 runs.


\begin{table}
\caption{The information of TCGA dataset. The tumor type categorization, tumor subtypes (primary diagnosis), ID of each primary diagnosis, and the number of test samples for each tumor are presented. }
\scriptsize 
\setlength{\tabcolsep}{.3pt}
\centering
\begin{tabular}{cccc}
\hline
\textbf{\begin{tabular}[c]{@{}c@{}}Tumor\\ Type\end{tabular}} & \textbf{Subtype}                                                                                              & \textbf{ID} & \textbf{\#Test Samples} \\ \hline
\multirow{2}{*}{\textbf{Brain}}                               & Brain Lower Grade Glioma                                                                                      & LGG         & 35                   \\
                                                              & Glioblastoma Multiforme                                                                                       & GBM         & 39                   \\ \hline
\multirow{3}{*}{\textbf{Endocrine}}                           & Adrenocortical Carcinoma                                                                                      & ACC         & 6                    \\
                                                              & Pheochromocytoma and   Paraganglioma                                                                          & PCPG        & 15                   \\
                                                              & Thyroid Carcinoma                                                                                             & THCA        & 51                   \\ \hline
\multirow{4}{*}{\textbf{Gastrointestinal}}                             & Colon Adenocarcinoma                                                                                          & COAD        & 33                   \\
                                                              & Rectum Adenocarcinoma                                                                                         & READ        & 11                   \\
                                                              & Esophageal Carcinoma                                                                                          & ESCA        & 14                   \\
                                                              & Stomach Adenocarcinoma                                                                                        & STAD        & 30                   \\ \hline
\multirow{3}{*}{\textbf{Gynecological}}                            & \begin{tabular}[c]{@{}c@{}}Cervical Squamous Cell Carcinoma \\ and Endocervical   Adenocarcinoma\end{tabular} & CESC        & 17                   \\
                                                              & Ovarian Serous   Cystadenocarcinoma                                                                           & OV          & 10                   \\
                                                              & Uterine Carcinosarcoma                                                                                        & UCS         & 3                    \\ \hline
\multirow{3}{*}{\textbf{Liver}}                               & Cholangiocarcinoma                                                                                            & CHOL        & 4                    \\
                                                              & Liver Hepatocellular Carcinoma                                                                                & LIHC        & 35                   \\
                                                              & Pancreatic Adenocarcinoma                                                                                     & PAAD        & 12                   \\ \hline
\multirow{2}{*}{\textbf{Mesenchymal}}                         & Uveal Melanoma                                                                                                & UVM         & 4                    \\
                                                              & Skin Cutaneous Melanoma                                                                                       & SKCM        & 24                   \\ \hline
\multirow{2}{*}{\textbf{Prostate/Testis}}                           & Prostate Adenocarcinoma                                                                                       & PRAD        & 40                   \\
                                                              & Testicular Germ Cell Tumors                                                                                   & TGCT        & 13                   \\ \hline
\multirow{3}{*}{\textbf{Pulmonary}}                           & Lung Adenocarcinoma                                                                                           & LUAD        & 43                   \\
                                                              & Lung Squamous Cell Carcinoma                                                                                  & LUSC        & 38                   \\
                                                              & Mesothelioma                                                                                                  & MESO        & 5                    \\ \hline
\multirow{4}{*}{\textbf{Urinary tract}}                       & Bladder Urothelial Carcinoma                                                                                  & BLCA        & 34                   \\
                                                              & Kidney Chromophobe                                                                                            & KICH        & 11                   \\
                                                              & Kidney Renal Clear Cell   Carcinoma                                                                           & KIRC        & 50                   \\
                                                              & Kidney Renal Papillary Cell   Carcinoma                                                                       & KIRP        & 28                   \\ \hline
\end{tabular}
\label{Data}
\end{table}

\begin{table}
\caption{The results of the search by EFV. The comparison between the efficiency of EFV, MFV, and KFV  is provided in terms of F1-scores (\%) by $K$-nearest neighbor approach. Whereas the size of   MFV is 1024 and the  size of EFV for each category is provided in the last column.   }
\scriptsize 
\setlength{\tabcolsep}{2pt}
\centering


\begin{tabular}{ccccccc}
\hline
\multirow{2}{*}{\textbf{Tumor Type}}       & \multicolumn{1}{l}{} & \multicolumn{3}{c}{\textbf{F1\_Score(\%)}}                                                             & \multicolumn{1}{l}{}                                                & \multirow{2}{*}{\textbf{\begin{tabular}[c]{@{}c@{}}Size of\\ EFV\end{tabular}}} \\ \cline{2-6}
                                           & \textbf{ID}          & \textbf{\begin{tabular}[c]{@{}c@{}}KFV\\  (\textasciitilde135*1024)\end{tabular}}    & \textbf{\begin{tabular}[c]{@{}c@{}}MFV \\ (1024)\end{tabular}} & \textbf{EFV}    & \textbf{\begin{tabular}[c]{@{}c@{}}Diff. \\ (EFV-KFV)\end{tabular}} &                                                                                  \\ \hline
\multirow{2}{*}{\textbf{Brain}}            & LGG                  & 81.08           & 84.51                                                          & \textbf{91.56}  & \textbf{10.48}                                                      & \multirow{2}{*}{\textbf{11}}                                                  \\
                                           & GBM                  & 81.08           & 85.71                                                          & \textbf{90.59}  & \textbf{9.51}                                                       &                                                                                  \\ \hline
\multirow{3}{*}{\textbf{Endocrine}}        & ACC                  & 44.41           & 54.55                                                          & \textbf{83.45}  & \textbf{39.04}                                                      & \multirow{3}{*}{\textbf{22}}                                                  \\
                                           & PCPG                 & 84.85           & 83.87                                                          & \textbf{92.47}  & \textbf{7.62}                                                       &                                                                                  \\
                                           & THCA                 & \textbf{100} & \textbf{100}                                                & 99.04           & -0.96                                                               &                                                                                  \\ \hline
\multirow{4}{*}{\textbf{Gastrointestinal}} & COAD                 & 76.47           & \textbf{76.71}                                                 & 76.52           & \textbf{0.05}                                                       & \multirow{4}{*}{\textbf{13}}                                                  \\
                                           & READ                 & 30           & \textbf{42.11}                                                 & 39.18           & \textbf{9.18}                                                       &                                                                                  \\
                                           & ESCA                 & 78.26           & \textbf{84.62}                                                 & 74.24           & -4.02                                                               &                                                                                  \\
                                           & STAD                 & \textbf{86.15}  & 79.31                                                          & 82.21          & -3.94                                                               &                                                                                  \\ \hline
\multirow{3}{*}{\textbf{Gynecological}}    & CESC                 & 94.12           & \textbf{97.14}                                                 & 95.53           & \textbf{1.41}                                                       & \multirow{3}{*}{\textbf{28}}                                                 \\
                                           & OV                   & \textbf{94.74}  & \textbf{94.74}                                                 & 92.76           & -1.98                                                               &                                                                                  \\
                                           & UCS                  & 85.71           & \textbf{100}                                                & 93.07           & \textbf{7.36}                                                       &                                                                                  \\ \hline
\multirow{3}{*}{\textbf{Hepatopancreato.}} & CHOL                 & 40           & 50                                                          & \textbf{64.01}  & \textbf{24.01}                                                      & \multirow{3}{*}{\textbf{12}}                                                  \\
                                           & LIHC                 & \textbf{95.77}  & \textbf{95.77}                                                 & 95.69           & -0.08                                                               &                                                                                  \\
                                           & PAAD                 & 76.19           & 73.68                                                          & \textbf{89.04}  & \textbf{12.85}                                                      &                                                                                  \\ \hline
\multirow{2}{*}{\textbf{Mesenchymal}}      & UVM                  & 66.67           & 66.67                                                          & \textbf{81.97}  & \textbf{15.30}                                                      & \multirow{2}{*}{\textbf{22}}                                                  \\
                                           & SKCM                 & 93.62           & 96                                                          & \textbf{97.52}  & \textbf{3.90}                                                       &                                                                                  \\ \hline
\multirow{2}{*}{\textbf{Prostate/Testis}}  & PRAD                 & \textbf{100} & \textbf{100}                                                & \textbf{100} & \textbf{0}                                                       & \multirow{2}{*}{\textbf{6}}                                                   \\
                                           & TGCT                 & \textbf{100} & \textbf{100}                                                & \textbf{100} & \textbf{0}                                                       &                                                                                  \\ \hline
\multirow{3}{*}{\textbf{Pulmonary}}        & LUAD                 & 78.33           & 81.58                                                          & \textbf{84.96}  & \textbf{6.63}                                                       & \multirow{3}{*}{\textbf{13}}                                                  \\
                                           & LUSC                 & 84.44           & 86.36                                                          & \textbf{87.44}  & \textbf{3}                                                       &                                                                                  \\
                                           & MESO                 & 75           & 75                                                          & \textbf{93.10}  & \textbf{18.10}                                                      &                                                                                  \\ \hline
\multirow{4}{*}{\textbf{Urinary tract}}    & BLCA                 & 93.15           & \textbf{94.29}                                                 & 93.32           & \textbf{0.17}                                                       & \multirow{4}{*}{\textbf{16}}                                                  \\
                                           & KICH                 & 85.71           & 90                                                          & \textbf{90.84}  & \textbf{5.13}                                                       &                                                                                  \\
                                           & KIRC                 & \textbf{96.97}  & 95.05                                                          & 95.62           & -1.35                                                               &                                                                                  \\
                                           & KIRP                 & \textbf{90.57}  & 86.27                                                          & 87.18           & -3.39                                                               &                                                                                  \\ \hline
\multicolumn{2}{c}{\textbf{Average}}                              & 81.28  &83.61                                                & \textbf{87.36}  & 6.08                                                      & \multicolumn{1}{c}{16}                                               \\ \hline
\multicolumn{2}{c}{\textbf{Std.}}                                 & 17.90  & 15.46                                                & 12.70  & 9.63                                                       & \multicolumn{1}{c}{6.45}                                                             \\ \hline
\multicolumn{2}{c}{\textbf{P-value}}                              &  &               0.033                                &0.004    &                                                       &                                             \\ \hline
\end{tabular}

\label{CFS}
\end{table}

\begin{figure*} 
\centering
\subfloat[Brain]{\includegraphics[width=0.8\textwidth]{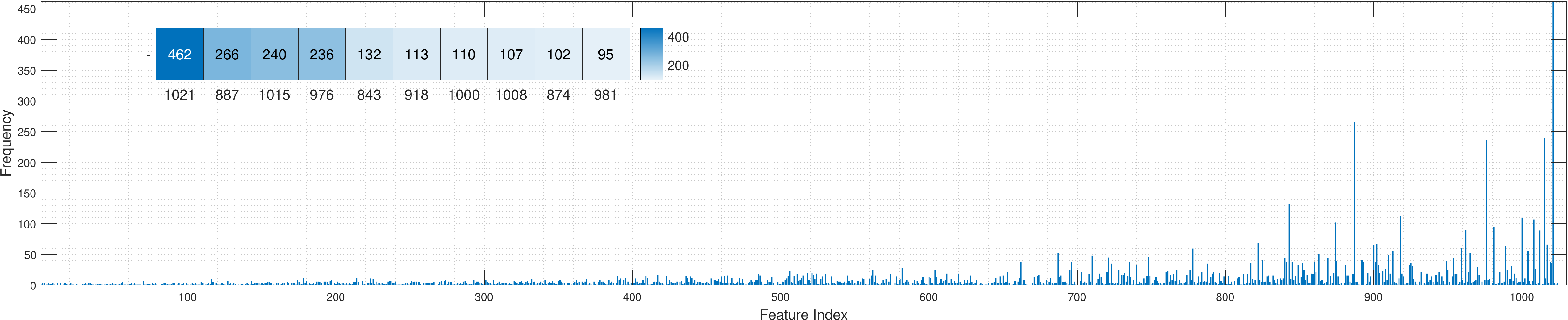}}
\\
\subfloat[Gastrointestinal]{\includegraphics[width=0.8\textwidth, height=80pt]{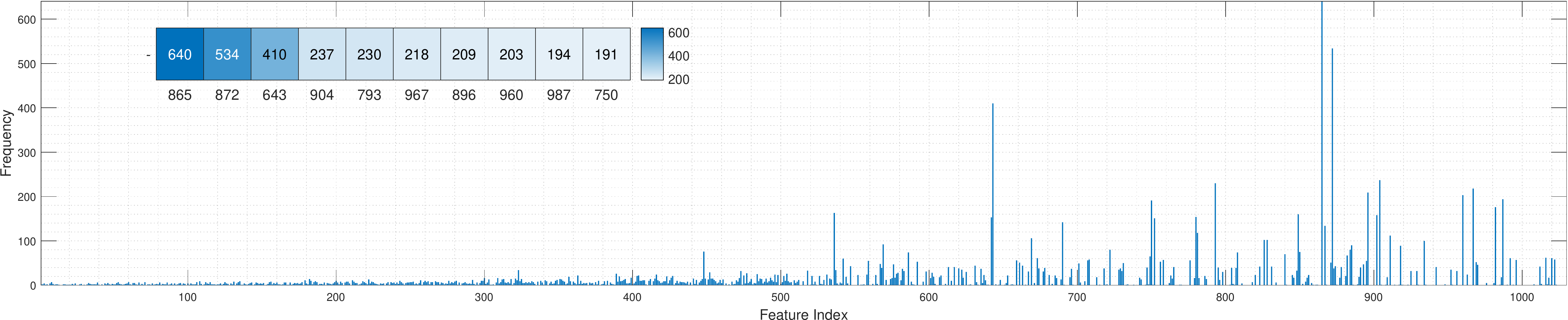}}
\\
\subfloat[Liver]{\label{fig:gull}\includegraphics[width=0.8\textwidth]{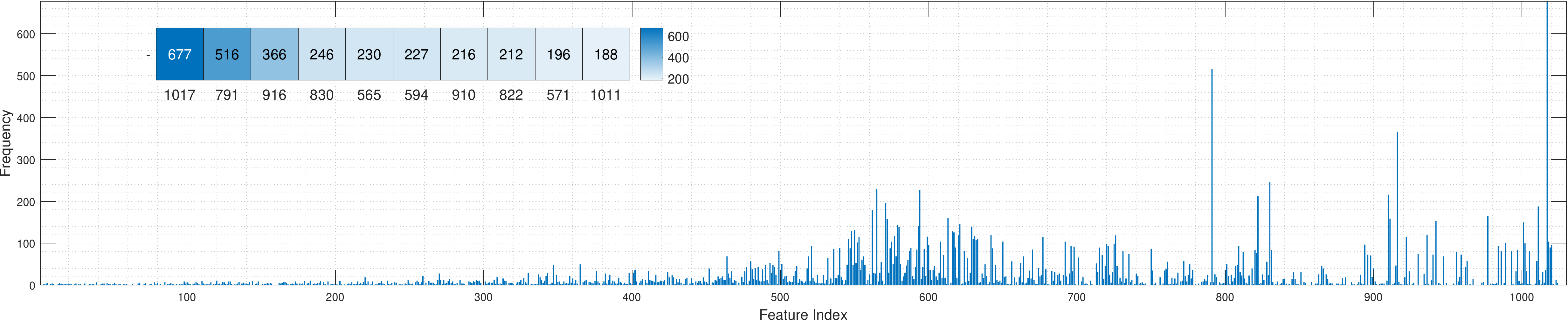}}
\\
\subfloat[Pulmonary]{\includegraphics[width=0.8\textwidth]{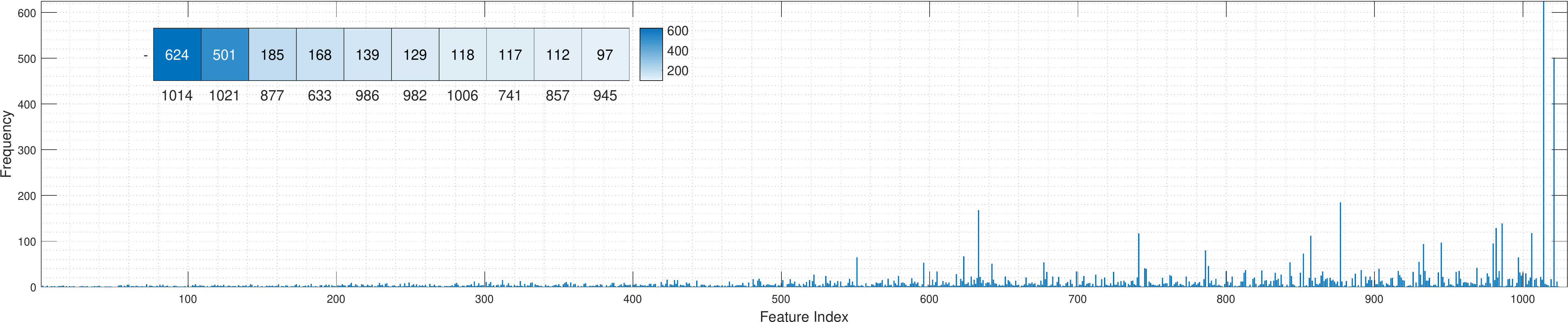}}
\caption{\small{Frequency of DNN-derived features for Brain, Gastrointestinal, Liver, and  Pulmonary sites. The 10  frequent features are presented by heatmap plot in which the value of frequency is showed inside the bin. The feature index is also presented in bottom of each plot. }}
\label{KimaHist}
\end{figure*}
\subsection{Effectiveness of Coarse Multi-objective Feature Selection}
In this section, we investigate the performance of the selected features by coarse optimization in terms of their accuracy in image retrieval. The results of the image search using the EFVs are compared with the results of the original features extracted by KimiaNet (i.e., KFV) and the results of
the MFVs (i.e., 1024 features). Table~\ref{CFS} represents the F1-score and the size of code for each method. The represented result of the EFV is the average value of the F1-score over the solutions of 31 runs. From each run, the feature subset with the maximum F1-scores is chosen. The size of the EFV is the average number of the selected features from the 31 runs. The last rows of the table indicate the mean and standard deviation (SD) of the F1-score over all tumor types along with the P value of the Wilcoxon test on the results of the MFV and EFV compared with that of KFV.\\
As Table~\ref{CFS} shows, the MFVs mostly surpass the original features, while they consist of a lower number of features. On average, there is an approximately 4\% improvement when using MFVs. Furthermore, the comparison between the results of the EFV and two other feature subsets reveals that feature selection could remarkably improve the results of the MFV and KFV methods.
The average F1-score resulting from feature selection is 87.36\%, which is ~4\% higher than the result of the image search using the MFV. Moreover, the average size of the EFV is 16.02, which is $1024/16.02=63.92$ times more compact. The improvement in the result of the EFVs compared to the result of the KFV method is almost 6\%. For some cancer subtypes, the difference between the results is significant. For instance, using evolutionary features achieved an accuracy of 83.45\% in the ACC  image search, while the resulting accuracy using the MFV was a much lower value of 54.55\%. The same level of improvement can be observed for the UVM and PADD cases. Overall, in 11 out of 26 cases, the EFV outperformed the other methods, and in 2 cases, the MFV or KFV obtained better results. In 13 cases, the differences between the two methods were not significant.

\begin{table*}
\caption{A comparison between the search results obtained using the 10  frequent features and  all EFVs on Pareto fronts. The number of features on  Pareto fronts is presented in the second column. For each tumor type and each feature, the  first value reflects that the corresponding feature from what percent of the evolutionary features is better. The ranks higher than 0.5 (i.e., the frequent feature is better than 50\% of  all features on Pareto front) are highlighted in bold. Furthermore, the difference between both results is presented as the second value. }
\scriptsize 
\centering
\begin{tabular}{lcc|cccccccccc|l}
\hline
\textbf{\begin{tabular}[c]{@{}c@{}}Tumor\\ Type\end{tabular}} & \textbf{\#Features}  & \textbf{} & \textbf{F1} & \textbf{F2} & \textbf{F3} & \textbf{F4} & \textbf{F5} & \textbf{F6} & \textbf{F7} & \textbf{F8} & \textbf{F9} & \textbf{F10} & \multicolumn{1}{c}{\textbf{Average}} \\ \hline
\multirow{2}{*}{\textbf{Brain}}                               & \multirow{2}{*}{238} & Rank      & 0.91        & 1        & 1        & 0.94        & 0.89        & 0.58        & 0.92        & 0.05        & 0.99        & 0.98        & \textbf{ 0.83}                                \\
                                                              &                      & Diff.     & 15.86       & 28.07       & 28.03       & 18.48       & 14.33       & 2.15        & 17.24       & -17.95      & 25.24       & 21.32        & 15.28                                \\ \hline
\multirow{2}{*}{\textbf{Endocrine}}                           & \multirow{2}{*}{507} & Rank      & 0.61        & 0.94        & 0.82        & 0.92        & 0.67        & 0.29        & 0.81        & 0.97        & 0.56        & 0.62        & \textbf{ 0.72}                                \\
                                                              &                      & Diff.     & 2.84        & 22.89       & 12.07       & 18.78       & 5.03        & -8.11       & 11.26       & 27.59       & 1.33        & 3.87         & 9.76                                 \\ \hline
\multirow{2}{*}{\textbf{Gastrointestinal}}                             & \multirow{2}{*}{177} & Rank      & 0.75        & 0.97        & 0.46        & 0.97        & 0.15        & 0.71        & 0.94        & 0.99        & 0.87        & 0.85       & \textbf{ 0.77}                                \\
                                                              &                      & Diff.     & 4.80        & 15.27       & -0.65       & 15.14       & -8       & 3.42        & 14.16       & 20.89       & 8.61        & 7.59         & 8.13                                 \\ \hline
\multirow{2}{*}{\textbf{Gynecological}}                             & \multirow{2}{*}{486} & Rank      & 0.69        & 0.66        & 0.50        & 0.31        & 0.08        & 0.99        & 0.82        & 0.23        & 0.12        & 0.77         & \textbf{0.52}                                 \\
                                                              &                      & Diff.     & 6.49        & 5.47        & 0.28        & -6.94       & -17.96      & 33.79       & 10.69       & -10.83      & -15.38      & 9.11         & 1.47                                 \\ \hline
\multirow{2}{*}{\textbf{Liver}}                               & \multirow{2}{*}{285} & Rank      & 0.84        & 0.74        & 0.94        & 0.32        & 0.95        & 0.87        & 0.65        & 0.73        & 0.77        & 0.99         & \textbf{0.78}                                 \\
                                                              &                      & Diff.     & 11.18       & 6.69        & 19.20       & -5.83       & 19.82       & 12.42       & 3.33        & 6.09        & 7.85        & 32.85        & 11.36                                \\ \hline
\multirow{2}{*}{\textbf{Melanocytic}}                         & \multirow{2}{*}{457} & Rank      & 0.91        & 0.04        & 0.60        & 0.36        & 0.36        & 0.88        & 0.91        & 0.36        & 0.14        & 0.36         & 0.49                                 \\
                                                              &                      & Diff.     & 26.37       & -6.28       & -2.98       & -4.04       & -4.04       & 13.53       & 26.37       & -4.04       & -5.14       & -4.04        & 3.57                                 \\ \hline
\multirow{2}{*}{\textbf{Prostate/Testis}}                    & \multirow{2}{*}{482} & Rank      & 0.78        & 0.97        & 0.77        & 0.94        & 0.83        & 0.90        & 0.70        & 0.67        & 0.65        & 0.82         & \textbf{0.80}                                \\
                                                              &                      & Diff.     & 18.20       & 31.01       & 16.28       & 30.87       & 22.02       & 28.10       & 11.95       & 9.93        & 8.01        & 21.07        & 19.74                                \\ \hline
\multirow{2}{*}{\textbf{Pulmonary}}                           & \multirow{2}{*}{222} & Rank      & 0.95        & 0.92        & 0.77        & 0.85        & 0.88        & 0.92        & 1        & 0.78        & 0.54        & 0.87         & \textbf{0.85}                                \\
                                                              &                      & Diff.     & 22.97       & 18.61       & 5.29        & 11.43       & 15.66       & 19.41       & 38.99       & 7.26        & -1.72       & 13.39        & 15.13                                \\ \hline
\multirow{2}{*}{\textbf{Urinary tract}}                       & \multirow{2}{*}{195} & Rank      & 0.97        & 0.91        & 0.91        & 0.92        & 0.72        & 0.82        & 0.84        & 0.94        & 1        & 0.97         & \textbf{.90}                                 \\
                                                              &                      & Diff.     & 21.28       & 15.72       & 15.50       & 16.20       & 6.31        & 10.28       & 12.26       & 17.24       & 27.42       & 20.85        & 16.31                                \\ \hline
\end{tabular}
\label{Rank2}
\end{table*}

\subsection{Effectiveness of the Frequent Features Histogram (FFH)}
 The main goal of this section is to investigate how the FFH can generate a more promising search space in which more robust solutions can be found. Fig.~\ref{KimaHist} shows the FFH of 1024 features resulting from several runs of the evolutionary algorithm for some tumor types. Apparently, despite exploring a very large search space, some features are frequently selected by the optimizer in different runs. For example, to accurately recognize the primary diagnosis for a brain image, feature 1021 is selected frequently by the optimizer as a discriminative feature. In some cases, a particular region of the histogram (i.e., indexes > 500) has higher density, which reveals that a specific portion of the network output is more beneficial for classification of the related tumor type. This can also be observed from the resultant features of the network so that the low-index features have mostly similar ranges of values. Therefore, they cannot be sufficiently discriminative for classifying WSIs, and consequently, they will be removed by the optimizer. On each plot, the indices of 10 more frequent features with the frequency values can also be observed with the heatmap plots.\\
Frequent features refer to features with the highest score calculated by Eq.~\ref{FR} captured in the FFH. To evaluate the quality of the frequent features, the features are compared with the set of features resulting from coarse optimization to obtain the best solution. Accordingly, the most frequent features from the sorted FFH (i.e., the most frequent features) are compared with the features on the Pareto front in terms of image search results. Each frequent feature is employed to find the three most similar WSIs for an image query. The F1-score is then calculated over all images. The same process is applied for each feature in the subsets on the Pareto front to compute the average F1-score over all features.
Table~\ref{Rank2} illustrates the results of the image search using two sets. For each tumor category, two rows are represented in the table; the first row shows the rank of the corresponding frequent feature compared to that of all features on the Pareto front. For instance, on brain tumor images, the most frequent feature (i.e., $F1$) is better than 91\% of 238 features on the Pareto front. Additionally, the second row indicates how much the search result obtained by the frequent features is different from that obtained by the EFVs. For example, the top feature in the list (i.e., $F1$) yields a 15.86\% higher accuracy than the average F1-score over the EFVs. According to the last column of the table, apart from the melanocytic and gynecological categories, the difference between the results of the two sets is higher than 77\% on all other tumor categories. In the urinary tract category, the top frequent features provide better results than 90\% of 195 features included in the solutions of the Pareto front.


\subsection{Effectiveness of Fine Multi-Objective Feature Selection }
Fine optimization is conducted on the frequent features captured in the FFH. The results of an image search using FFVs are obtained and evaluated accordingly. The components of the multi-objective  algorithm are the same as those of coarse optimization; however, the search space has been shrunk. Since the dimension of the problem is decreased significantly (i.e., from 1024D to 30D), the computational complexity of fine-level optimization is greatly reduced. In this section, the quality of features selected by fine optimization is assessed in terms of image search accuracy and stability.

\begin{table}
\caption{A comparison between the results using FFV and EFV resulted by the fine and coarse optimization, respectively. The number of features and the  F1-score resulted from both optimization process are represented. 
The last two columns shows the result by ordered selection from FFH with highest score which is validated by Wilcoxon statistical test. The  number of selected frequent features is considered equally to the size of FFV. }
\scriptsize 
\setlength{\tabcolsep}{3pt}
\centering
\begin{tabular}{cccccccc}
\hline
\multirow{2}{*}{\textbf{Tumor Type}}    & \multirow{2}{*}{\textbf{ID}} & \multicolumn{2}{c}{\textbf{FFV}} & \multicolumn{2}{c}{\textbf{EFV}} & \multicolumn{2}{c}{\textbf{\begin{tabular}[c]{@{}c@{}}Ordered \\ Selection\end{tabular}}} \\ \cline{3-8} 
                                        &                              & \textbf{\begin{tabular}[c]{@{}c@{}}F1-Score\\ (\%)\end{tabular}}                          & \textbf{\# F}                                & \textbf{\begin{tabular}[c]{@{}c@{}}F1-Score\\ (\%)\end{tabular}}                           & \textbf{\# F}                                 & \textbf{\begin{tabular}[c]{@{}c@{}}F1-Score\\ (\%)\end{tabular}}                          & \textbf{\# F}                                \\ \hline
\multirow{2}{*}{\textbf{Brain}}         & LGG                          & \textbf{94.11}                             & \multirow{2}{*}{4}                           & 91.56                                       & \multirow{2}{*}{11}                           & 82.13                                      & \multirow{2}{*}{4}                           \\
                                        & GBM                          & \textbf{93.46}                             &                                              & 90.59                                       &                                               & 78.85                                      &                                              \\ \hline
\multirow{3}{*}{\textbf{Endocrine}}     & ACC                          & \textbf{93.88}                             & \multirow{3}{*}{12}                          & 83.45                                       & \multirow{3}{*}{22}                           & 54.40                                      & \multirow{3}{*}{12}                          \\
                                        & PCPG                         & \textbf{95.99}                             &                                              & 92.47                                       &                                               & 83.88                                      &                                              \\
                                        & THCA                         & \textbf{99.60 }                                    &                                              & \textbf{99.04 }                                    &                                               & \textbf{99.76}                             &                                              \\ \hline
\multirow{4}{*}{\textbf{Gastro.}}       & COAD                         & \textbf{79.17}                             & \multirow{4}{*}{9}                           & 76.52                                       & \multirow{4}{*}{13}                           & 75.72                                      & \multirow{4}{*}{9}                           \\
                                        & READ                         & \textbf{45.19}                             &                                              & 39.18                                       &                                               & 25.28                                      &                                              \\
                                        & ESCA                         & \textbf{82.45}                             &                                              & 74.24                                       &                                               & 73.81                                      &                                              \\
                                        & STAD                         & \textbf{82.24}                             &                                              & \textbf{82.21}                                       &                                               & 80.17                                      &                                              \\ \hline
\multirow{3}{*}{\textbf{Gynaeco.}}      & CESC                         & \textbf{94.87 }                                     & \multirow{3}{*}{16}                          & \textbf{95.53}                              & \multirow{3}{*}{28}                           & 88.21                                      & \multirow{3}{*}{16}                          \\
                                        & OV                           & \textbf{89.96}                                      &                                              & \textbf{92.76}                              &                                               & 77.82                                      &                                              \\
                                        & UCS                          & \textbf{100}                            &                                              & 93.07                                       &                                               & 95                                      &                                              \\ \hline
\multirow{3}{*}{\textbf{Liver}}         & CHOL                         & \textbf{75.34}                             & \multirow{3}{*}{9}                           & 64.01                                       & \multirow{3}{*}{12}                           & 20.29                                      & \multirow{3}{*}{9}                           \\
                                        & LIHC                         & \textbf{95.46}                             &                                              & \textbf{95.69}                                      &                                               & 94.22                                      &                                              \\
                                        & PAAD                         & \textbf{91.17}                             &                                              & \textbf{89.04}                                       &                                               & 84.15                                      &                                              \\ \hline
\multirow{2}{*}{\textbf{Mesenchymal}}   & UVM                          & 66.67                                      & \multirow{2}{*}{8}                           & \textbf{81.97}                              & \multirow{2}{*}{22}                           & 65.08                                      & \multirow{2}{*}{8}                           \\
                                        & SKCM                         & 96                                      &                                              & \textbf{97.52}                              &                                               & 95.65                                      &                                              \\ \hline
\multirow{2}{*}{\textbf{Prostrate}}     & PRAD                         & \textbf{100}                            & \multirow{2}{*}{14}                          & \textbf{100}                             & \multirow{2}{*}{6}                            & 99.38                                      & \multirow{2}{*}{14}                          \\
                                        & TGCT                         & \textbf{100}                            &                                              & \textbf{100}                             &                                               & 98.08                                      &                                              \\ \hline
\multirow{3}{*}{\textbf{Pulmonary}}     & LUAD                         & \textbf{86.05}                             & \multirow{3}{*}{8}                           & \textbf{84.96}                                       & \multirow{3}{*}{13}                           & \textbf{84.51}                                      & \multirow{3}{*}{8}                           \\
                                        & LUSC                         & \textbf{88.61}                             &                                              & \textbf{87.44}                                       &                                               & 86.39                                      &                                              \\
                                        & MESO                         & \textbf{98.15}                             &                                              & 93.10                                       &                                               & 76.65                                      &                                              \\ \hline
\multirow{4}{*}{\textbf{Urinary tract}} & BLCA                         & \textbf{93.28}                             & \multirow{4}{*}{9}                           & \textbf{93.32}                                       & \multirow{4}{*}{16}                           & 91.96                                      & \multirow{4}{*}{9}                           \\
                                        & KICH                         & \textbf{94.51}                             &                                              & 90.84                                       &                                               & 83.20                                      &                                              \\
                                        & KIRC                         & \textbf{95.91}                             &                                              & \textbf{95.62}                                       &                                               & 93.41                                      &                                              \\
                                        & KIRP                         & \textbf{86.15}                                      &                                              & \textbf{87.18}                              &                                               & 80.37                                      &                                              \\ \hline
\multicolumn{2}{c}{\textbf{Average}}                                   & \textbf{89.16}                             & \textbf{10}                                  & 87.36                                       & 16                                            & 79.55                                      & \textbf{10}                                  \\ \hline
\multicolumn{2}{c}{\textbf{Std.}}                                      & 11.95                                      & 3.38                                         & 12.70                                       & 6.45                                          & 19.38                                      & 3.38                                         \\ \hline
\multicolumn{2}{c}{\textbf{P-value}}                                      &                                      &                                          &                            0.016        &                                          &    4.57e-05                              &                                          \\ \hline
\end{tabular}
\label{Fine}
\end{table}

\subsubsection{Image Search Results using FFVs}
\begin{figure*}
\centering
\includegraphics[width=0.8\textwidth]{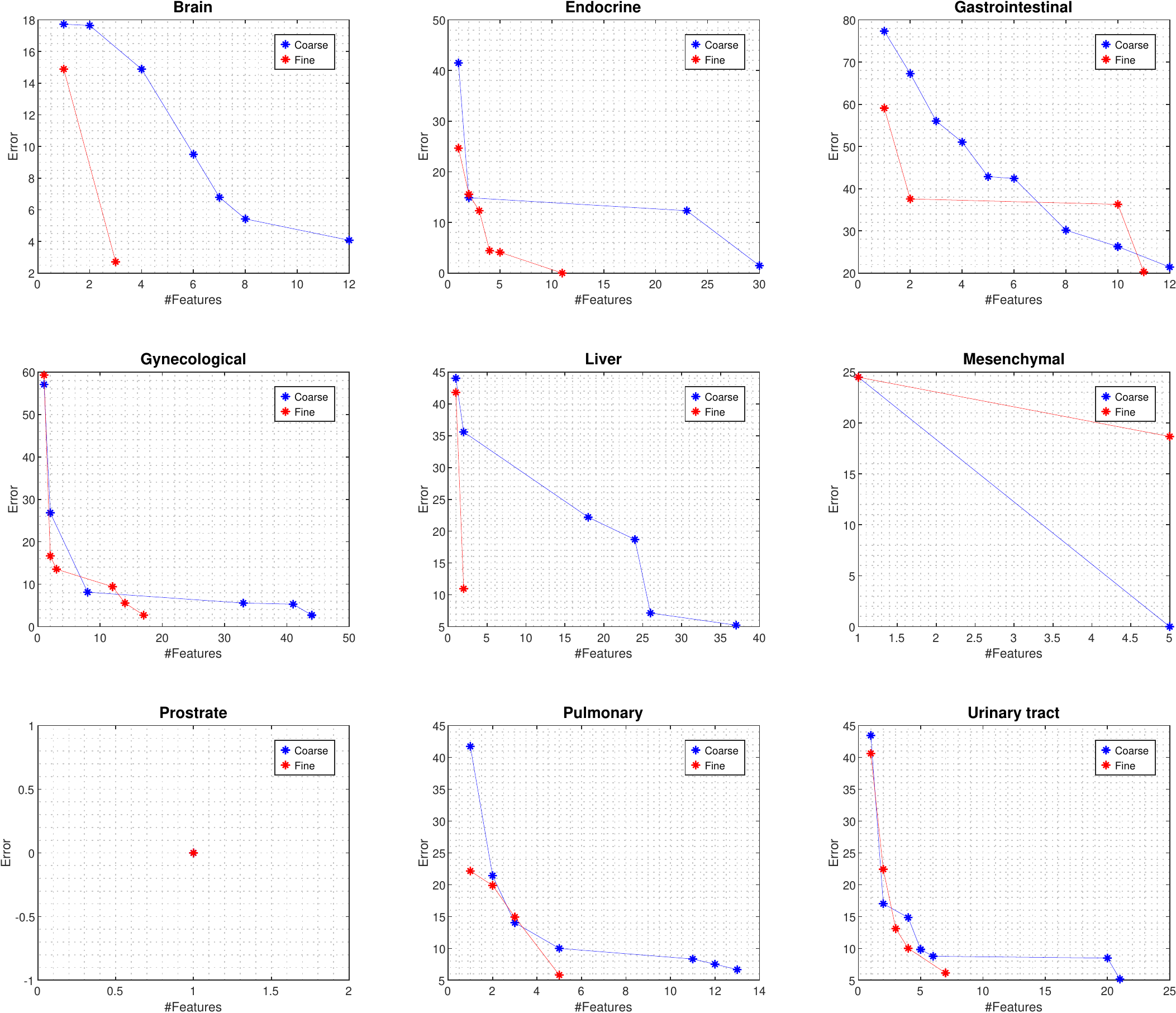}

\caption{A comparison on  Pareto fronts resulted from the fine and coarse optimization. The $x$-axis and $y$-axis indicate the number of selected features in each solution and the complement  value of the F1-score, respectively.   }
\label{Fig:PF}
\end{figure*}

\begin{figure*}
\centering
\includegraphics[width=1\textwidth]{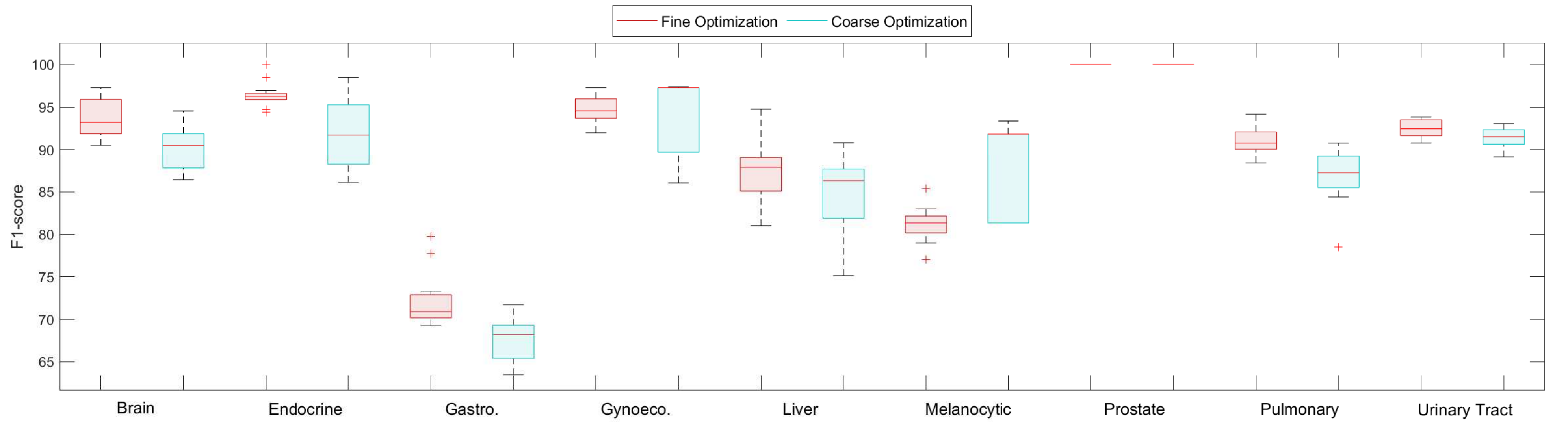}
\caption{The boxplot of the results using FFV and EFV. The red plots represents the F1-score of fine optimization and the blue plots shows the coarse optimization results.  }
\label{Fig:Boxplot}
\end{figure*}

The average F1-score over 31 Pareto fronts is presented in Table~\ref{Fine}. From each Pareto front, the subset with the maximum F1-score is selected. From the table, it is indicated that not only does fine optimization improve the results of the image search, but the number of selected features also decreases correspondingly. In fine optimization, the average number of features over the best subsets of the Pareto fronts is 10, whereas this number was 16 for coarse optimization. Moreover, on most of the tumor types, the F1-score is improved by the selected subset from the frequent features. Although the substantial goal of fine optimization is not necessarily to compete with coarse optimization, FFVs augment the accuracy of image retrieval. On the other hand, the lower number of features can augment the digital processing of pathology images by reaching fundamental goals;, that is, 1) smaller code provides fast retrieval among millions of images, 2) using the knowledge obtained from optimal features can lead to designing more compact and efficient DNNs, and 3) the demand for large amounts of memory to save descriptive data of biopsy samples decreases considerably . Therefore, although fine optimization is a postprocess of coarse optimization, it only needs to be ran once, and its long-life benefits will be advantageous when the software is operational in hospitals.\\
As previously shown, the frequent features have high quality and can be used to generate efficient subsets but finding the best combination is needed. To verify the effectiveness of the optimization of frequent features, we compare the feature subsets of fine optimization with those of coarse optimization and the ordered selection explained below.\\
One way that a feature subset can be generated using frequent features is to rank them according to their scores (calculated by Eq.\ref{FR}) and then select a subset from the top of the list. The number of selected features from the list is equal to the size of the best subset resulting from fine optimization.
By conducting this fair comparison, we show that finding the optimal combination based on optimization is a crucial challenge and that selection based on the frequency alone is not a sufficient condition, which justifies the vital role of the fine optimization stage. Table~\ref{Fine} shows the results of this comparison in terms of the F1-score. From the table, it is indicated that the difference between these two schemes is significant, as revealed by the p value presented in the last row. Furthermore, the average of the image retrieval results by the FFV is 89.16\% while an ordered selection of frequent features results in a 79.55\% F1-score. For all primary diagnoses except the THCA category, FFVs surpass the ordered high-frequency features. The highlighted values in the table represent the winning results, which are significantly different compared to those of the competitors , are specified based on the P value obtained for each tumor type. Overall, in 20 out of 26 cases, fine optimization outperformed the other methods, and in 6 cases, coarse optimization or ordered selection obtained better results.\\
An alternative indicator of the quality of FFVs is the final Pareto fronts (Fig.~\ref{Fig:PF}), which are the nondominated solutions remaining on the Pareto front. The Pareto fronts emerging from fine optimization result in subsets with a smaller number of features and lower retrieval error. For a min-min presentation of the Pareto fronts, the $y$-axis reflects the complement of the F1-score as the error, and the $x$-axis indicates the number of selected features in each solution. In some categories, such as the brain, endocrine, gastrointestinal, and gynecological categories, fine optimization obtains results closer to the optimal Pareto front. In other words, fine optimization subsets with lower errors and fewer features are obtained. For instance, the two resulting solutions obtained by fine optimization could dominate all solutions on the coarse Pareto front of the brain or endocrine categories with the same number of features and fewer errors. Consequently, the range of the selected features from the frequent features is smaller than the range of all features. On the prostate/testis category, since the features are sufficiently discriminative, only one feature is required to separate the tumor subtypes with an accuracy of 100\%.
In addition, the boxplots in Fig.~\ref{Fig:Boxplot} represent the comparison between the results of 31 runs of fine and coarse optimization. From the figure, the SD of the results on coarse optimization is significantly higher than that on fine optimization. Furthermore, the median F1-score resulting from 31 runs of fine optimization is higher than that from coarse optimization.
The difficulty of classification on different problems (i.e., tumor types) can be determined based on the resulting F1-scores; however, the dimension of the optimization problem is the same for all types. For instance, from the Table II, the primary diagnosis for the brain category can be discriminated with ~94\% accuracy using FFVs, while the gastrointestinal cancer types are classified with ~72\% accuracy.
This was expected because READ and COAD are similar neoplasms of different anatomical sites. These adenocarcinomas are usually morphologically identical. Thus, the average value of the F1-score for each tumor type can be defined as a measure of the difficulty of the problem. Accordingly, we can sort the problems considering their difficulty in descending order,
 as follows: gastrointestinal, mesenchymal, liver, pulmonary, gynecological, urinary tract, endocrine, brain and prostate.

\subsubsection{Evaluation of the Stability of FFVs}

 As mentioned previously, innovization supports the knowledge transfer from a set of stochastic runs of an evolutionary algorithm to obtain more robust solutions. Accordingly, generating a shrunken search space for fine optimization is expected to produce more reliable feature subsets compared to coarse optimization. Therefore, the robustness of the FFV and EFV is evaluated in terms of one of the well-known stability measures. Feature selection is often considered stable if the selected features are consistent through multiple runs of the algorithm. Thus, the stability is calculated by the amount of overlap between the subsets of selected features.
  In general, to compute the stability of the results of a feature selection algorithm, the average similarity over all pairwise feature subsets should be calculated using Eq.~\ref{S}.

 \begin {equation}
 \label {S}
S=  \frac 2 {L(L-1)} \sum _{i=1} ^{L} \sum _{j=i+1} ^{L} J(FS_i,FS_j),
\end {equation} 

where $L$ denotes the number of feature subsets resulting from $R$ runs of the algorithm and $J(FS_i,FS_j)$ is Jaccard's index~\cite{yang2013stability} between two feature subsets, namely, $FS_i$ and $FS_j$. The stability index ($S$) returns a number in the range of [0,1], where a value close to 0 indicates that the feature selection algorithm is unstable and a value close to 1 is indicative of a stable algorithm. Formally, this index is defined by Eq.~\ref{J}. 

 \begin {equation}
 \label {J}
 J(FS_i,FS_j)=\frac{|FS_i\cap FS_j|}{|FS_i\cup FS_j|}
\end {equation}

Jaccard's index calculates the ratio of the intersection between the selected features in two subsets over the accumulative number of all features in both subsets.
To assess the stability of the results, the measure is calculated over the feature subsets with the best F1-score from 31 independent runs of both algorithms, as presented in Table~\ref{Jaccard}. It can be observed from the table that the FFVs possess more common features than the EFVs, which leads to higher stability. However, in fine optimization, due to the reduction in dimensionality of the search space, it is transparently expected that the intersection between the resulting subsets increases. Moreover, the average value of the Jaccard index on fine optimization is 0.2775, whereas coarse optimization results in 0.0478 of this measure, which is ~6 times smaller.


\begin{table}[ht]
\caption{The values of Jaccard's index as a measure for feature selection stability; a higher value is better.}
\scriptsize 
\centering
\begin{tabular}{lcc}
\hline
\multicolumn{1}{l}{Tumor Type} & FFV & EFV \\ \hline
Brain                          & \textbf{0.2699}                                              & 0.0478                                                         \\
Endocrine                      & \textbf{0.3716}                                              & 0.0402                                                         \\
Gastrointestinal                        & \textbf{0.3226}                                              & 0.0483                                                         \\
Gynecological                        & \textbf{0.3585}                                              & 0.0848                                                         \\
Liver                          & \textbf{0.2448}                                              & 0.0495                                                         \\
Melanocytic                    & \textbf{0.2109}                                              & 0.0312                                                         \\
Prostate/Testis               & \textbf{0.2581}                                              & 0.0374                                                         \\
Pulmonary                      & \textbf{0.2207}                                              & 0.0590                                                         \\
Urinary tract                  & \textbf{0.2403}                                              & 0.0318                                                         \\ \hline
\textbf{Average}               & \textbf{0.2775}                                              & 0.0478                                                         \\ \hline
\end{tabular}
\label{Jaccard}
\end{table} 


\subsection{Effectiveness of the Selected Features in the Proposed  Decision  Space}
\begin{figure*}
\centering

\includegraphics[width=.8\textwidth]{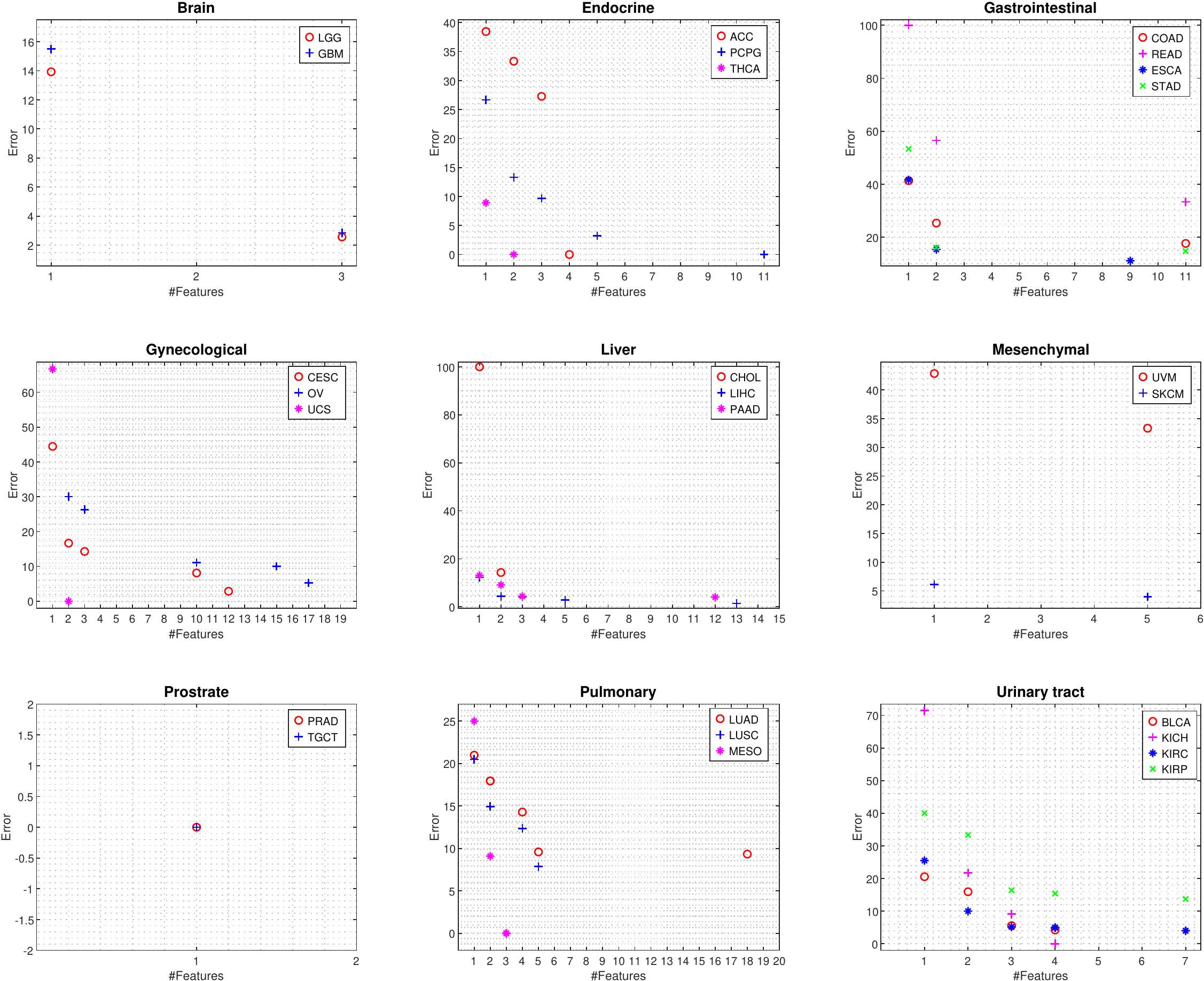}\\

 \caption{The Pareto fronts in decision space. Each plot illiterates the Pareto fronts of the solutions resulted for classification of each primary diagnosis. For instance for Liver, three Pareto fronts on the results of image retrieval for  CHOL, LIHC, and PAAD  are generated in different colors. }
\label{DM}
\end{figure*}

A multi-objective  optimization problem is made of M single-objective optimization problems. Moreover, each objective can be a combination of multiple embedded subobjectives. In this study, as mentioned previously, the quality of features is evaluated in terms of the average F1-score over all primary diagnoses of a tumor site. However, the efficiency of the selected features on all tumor subtypes inside a category is not equal. As Table~\ref{Fine} indicates, in some categories, the variance between the resulting F1-scores on distinct subtypes is significant. For instance, in the gastrointestinal category, the resulting F1-score for the READ class using the optimal features is 45.19\%, whereas this objective has a value higher than 79\% on other tumor subtypes belonging to this category. Although considering the average accuracy of different primary diagnoses aids the optimizer with fewer objectives and consequently forms a simplified optimization problem, this approach affects the decision-making process because of the lower transparency caused by embedded subobjectives. For each multi-objective optimization problem, decision making is a major complementary process that should be considered seriously. In this section, we define a paradigm by which the decision maker can transparently consider the subobjectives as well. For this purpose, a decision space is constructed from the last step of the multi-objective optimization process to represent the value of subobjectives separately. The obtained solutions can generate a low-level Pareto front for each primary diagnosis. To this end, the results of the image retrieval process using each subset on the Pareto front are obtained for each primary diagnosis separately, and thereafter, the nondominated solutions create the Pareto front for the corresponding primary diagnosis. In this way, the decision maker can effectively choose a subset of features according to the importance of the corresponding tumor subtype. Fig.~\ref{DM} represents the low-level Pareto fronts of the primary diagnosis. From the figure, the variance in the efficiency of the FFVs for different tumor subtypes can be observed. For instance, for the gastrointestinal type, a subset with 11 features can distinguish READ with a minimum error of 32\% whereas the minimum error on ESCA can be obtained using an FFV of size 2. These Pareto fronts provide the decision maker with more information about the newly defined decision space to effectively select one of the feature subsets according to the desired goal, such as determining  the importance of a specific tumor subtype. As another explicit instance, a different performance of the optimal features can be observed for each tumor subtype of the urinary tract category. For classification of BLCA, 3 features selected by the optimizer could decrease the error to 5\%, while the minimum value of the error on three other primary diagnoses in this category can be obtained using a subset of size 7. Depending on the desired compactness and the level of accuracy for a specific tumor subtype, a subset can be efficiently selected by an expert user.

\subsection{Feature Visualization}

\begin{figure}
\centering
\begin{tabular}{c}

      \includegraphics[width=0.90\linewidth]{Figures/feature_visualization/TCGA-DC-6156-01Z-00-DX1.pdf}\\ 
      (a) Agglomerative hierarchical clustering for READ\\
       \includegraphics[width=0.70\linewidth]{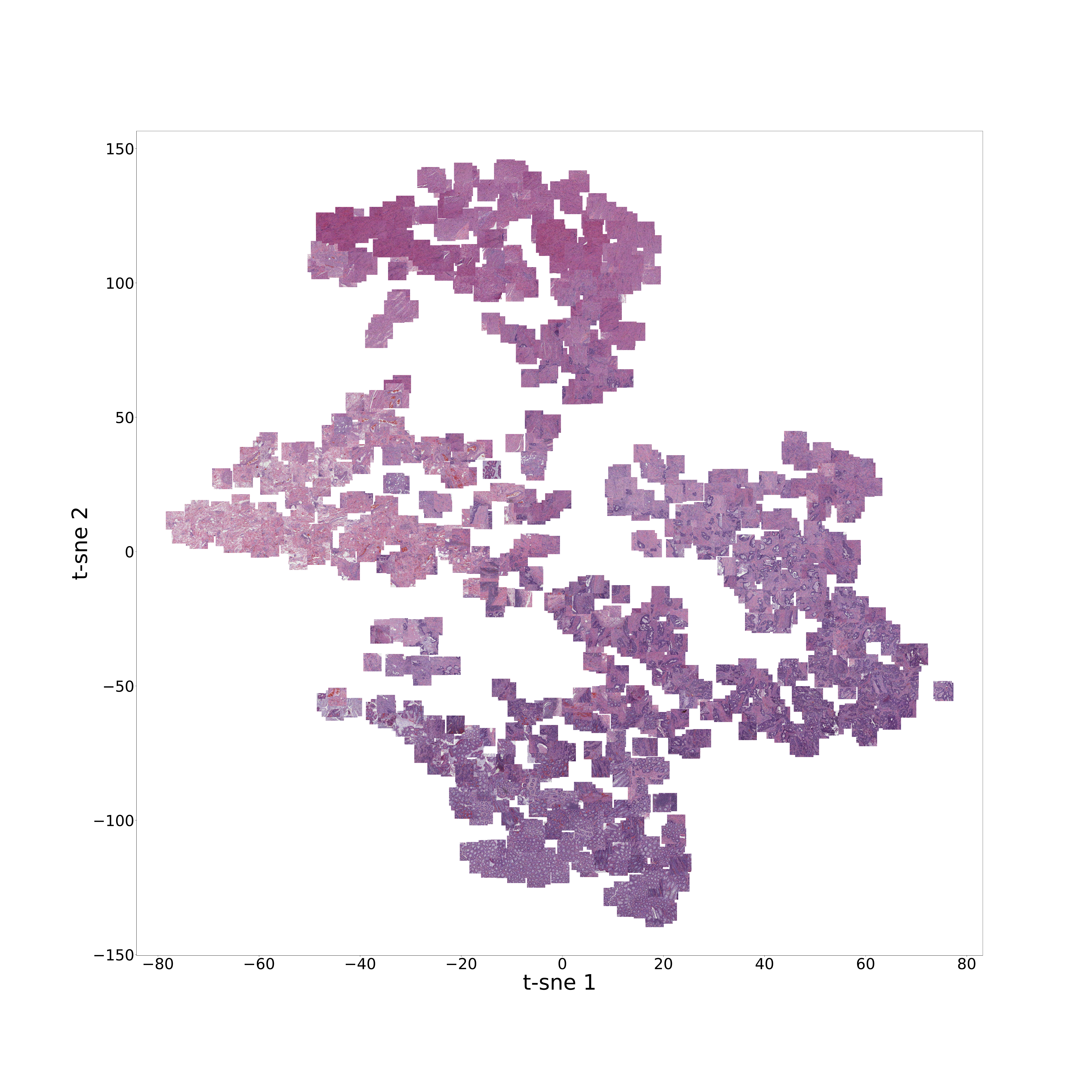}\\
       (b) t-SNE visualization for READ\\

\end{tabular}
\caption{a) Tissue patches of a WSI (rectum adenocarcinoma) are grouped into 4 clusters using agglomerative hierarchical clustering, and the gastrointestinal the evolutionary subset resulted by fine optimization. It shows that each cluster corresponds to a specific histologic pattern. b) t-SNE visualization of tissue patches based on their gastrointestinal optimized subset, one could see that tissue patches with similar visual patterns are standing close together. }
\label{fig: dfvis}
\end{figure}

The major reasons for decreasing the lengths of the deep feature vectors are to reduce the storage space required for tissue patch representation, to speed up the image search and to increase the retrieval accuracy. As per the previously explained results, it would be plausible to store 100 more patch representations per WSI if the deep feature vectors were of size 10 instead of size 1024. Consequently, this not only makes WSI representation more comprehensive but also helps hospitals and healthcare institutions decrease their digital storage costs. However, these optimized sets of deep features might lack the information required to represent noncancerous tissue, since they have been optimized for cancer subtype classification. We conducted two sets of experiments to evaluate whether tissue patches with similar DNN-derived feature vectors are visually similar. DNN-derived feature vectors were extracted from tissue patches of a WSI. For each vector, the evolutionary gastrointestinal features were selected. First, agglomerative hierarchical clustering \cite{scikit-learn} was applied to tissue patches to cluster them into 4 groups. Fig. \ref{fig: dfvis} shows that the clusters are selective for specific regions of the WSI. Additionally, 2D t-SNE visualization of the tissue patches demonstrates that those with similar visual patterns are close together, as shown in Fig. \ref{fig: dfvis} (b), as expected. These findings imply that optimized subsets contain adequate information to distinguish the different tissue types of a WSI, whether it represents malignant or nonmalignant tissue.

\section{Conclusions}

In this paper, we proposed a novel approach based on a novel hybridization of a DNN and an evolutionary computation for the optimal design of DNN embeddings. In fact, instead of the optimally designed deep networks, we proposed enhancing the results of a deep network by using two-stage robust DNN-derived feature selection based on the innovization concept. The second stage of feature selection (called fine feature optimization) is based on the FFH and dimensionality reduction, which results in a robust and effective feature selection result.
Processing a gigapixel pathology image for a task such as an image search is accomplished effectively by evolutionary computing and deep learning. The obtained results are highly promising and represent the gigapixel images with a compact and reliable DNN-derived feature vector suitable for fast and accurate CBIR. The proposed approach employed the consolidated power of deep learning, center-based sampling, search space and dimension reduction, and Darwinian innovization (i.e., survival of the fittest) to overcome the overwhelming complexity of digitized biopsy samples in histopathology. 
Additionally, in this study, a novel decision spfaknace is introduced for MOOs to facilitate expert decision-making processes by offering better transparency on embedded subobjectives. We established a 13,800-fold memory complexity reduction process (i.e., 13,800 features are reduced to 10 features on average to represent a WSI). 
In addition, the accuracy of classification was improved by approximately 8$\%$ compared with the state-of-the-art literature on the largest public dataset. The authors believe that the novel contributions achieved in this study, as a result of evolution in action, will play a crucial role in digital pathology.

\section{Acknowledgment}
This work is supported by Ontario Research Fund (ORF). 

\bibliographystyle{IEEEtran}
\bibliography{main.bib}

\begin{thebibliography}{10}
\providecommand{\url}[1]{#1}
\csname url@samestyle\endcsname
\providecommand{\newblock}{\relax}
\providecommand{\bibinfo}[2]{#2}
\providecommand{\BIBentrySTDinterwordspacing}{\spaceskip=0pt\relax}
\providecommand{\BIBentryALTinterwordstretchfactor}{4}
\providecommand{\BIBentryALTinterwordspacing}{\spaceskip=\fontdimen2\font plus
\BIBentryALTinterwordstretchfactor\fontdimen3\font minus
  \fontdimen4\font\relax}
\providecommand{\BIBforeignlanguage}[2]{{%
\expandafter\ifx\csname l@#1\endcsname\relax
\typeout{** WARNING: IEEEtran.bst: No hyphenation pattern has been}%
\typeout{** loaded for the language `#1'. Using the pattern for}%
\typeout{** the default language instead.}%
\else
\language=\csname l@#1\endcsname
\fi
#2}}
\providecommand{\BIBdecl}{\relax}
\BIBdecl

\bibitem{Doley}
M.~A. Dooley, C.~Aranow, and E.~M. Ginzler, ``Review of {ACR} renal criteria in
  systemic lupus erythematosus,'' \emph{Lupus}, vol.~13, no.~11, pp. 857--860,
  2004, pMID: 15580982.

\bibitem{brunt_2010}
E.~M. Brunt, ``Histopathology of nonalcoholic fatty liver disease,''
  \emph{World Journal of Gastroenterology}, vol.~16, no.~42, p. 5286, Nov 2010.

\bibitem{emre_bilge}
S.~Emre, I.~Bilge, A.~Sirin, I.~Kılıcaslan, A.~Nayır, F.~Oktem, and
  V.~Uysal, ``Lupus nephritis in children: Prognostic significance of
  clinicopathological findings,'' \emph{Nephron}, vol.~87, no.~2, p. 118–126,
  2001.

\bibitem{goodman_ishak}
Z.~Goodman and K.~Ishak, ``Histopathology of hepatitis c virus infection,''
  \emph{Seminars in Liver Disease}, vol.~15, no.~01, p. 70–81, Jan 1995.

\bibitem{Pantanowitz2015}
L.~Pantanowitz, N.~Farahani, and A.~Parwani, ``{Whole slide imaging in
  pathology: advantages, limitations, and emerging perspectives},''
  \emph{Pathology and Laboratory Medicine International}, no. July, p.~23,
  2015.

\bibitem{KALRA_yottixel}
S.~Kalra, H.~Tizhoosh, C.~Choi, S.~Shah, P.~Diamandis, C.~J. Campbell, and
  L.~Pantanowitz, ``Yottixel – an image search engine for large archives of
  histopathology whole slide images,'' \emph{Medical Image Analysis}, vol.~65,
  p. 101757, 2020.

\bibitem{Saritha2019}
R.~R. Saritha, V.~Paul, and P.~G. Kumar, ``Content based image retrieval using
  deep learning process,'' \emph{Cluster Computing}, vol.~22, no.~2, pp.
  4187--4200, Mar 2019.

\bibitem{xue2012particle}
B.~Xue, M.~Zhang, and W.~N. Browne, ``Particle swarm optimization for feature
  selection in classification: A multi-objective approach,'' \emph{IEEE
  transactions on cybernetics}, vol.~43, no.~6, pp. 1656--1671, 2012.

\bibitem{riasatian2021fine}
A.~Riasatian, M.~Babaie, D.~Maleki, S.~Kalra, M.~Valipour, S.~Hemati,
  M.~Zaveri, A.~Safarpoor, S.~Shafiei, M.~Afshari \emph{et~al.}, ``Fine-tuning
  and training of densenet for histopathology image representation using tcga
  diagnostic slides,'' \emph{Medical Image Analysis}, vol.~70, p. 102032, 2021.

\bibitem{gutman2013cancer}
D.~A. Gutman, J.~Cobb, D.~Somanna, Y.~Park, F.~Wang, T.~Kurc, J.~H. Saltz,
  D.~J. Brat, L.~A. Cooper, and J.~Kong, ``Cancer digital slide archive: an
  informatics resource to support integrated in silico analysis of tcga
  pathology data,'' \emph{Journal of the American Medical Informatics
  Association}, vol.~20, no.~6, pp. 1091--1098, 2013.

\bibitem{xue2015survey}
B.~Xue, M.~Zhang, W.~N. Browne, and X.~Yao, ``A survey on evolutionary
  computation approaches to feature selection,'' \emph{IEEE Transactions on
  Evolutionary Computation}, vol.~20, no.~4, pp. 606--626, 2015.

\bibitem{hosseini2019evolutionary}
E.~S. Hosseini and M.~H. Moattar, ``Evolutionary feature subsets selection
  based on interaction information for high dimensional imbalanced data
  classification,'' \emph{Applied Soft Computing}, vol.~82, p. 105581, 2019.

\bibitem{deb2016breaking}
K.~Deb and C.~Myburgh, ``Breaking the billion-variable barrier in real-world
  optimization using a customized evolutionary algorithm,'' in
  \emph{Proceedings of the Genetic and Evolutionary Computation Conference
  2016}, 2016, pp. 653--660.

\bibitem{pes2017exploiting}
B.~Pes, N.~Dess{\`\i}, and M.~Angioni, ``Exploiting the ensemble paradigm for
  stable feature selection: a case study on high-dimensional genomic data,''
  \emph{Information Fusion}, vol.~35, pp. 132--147, 2017.

\bibitem{ambroise2002selection}
C.~Ambroise and G.~J. McLachlan, ``Selection bias in gene extraction on the
  basis of microarray gene-expression data,'' \emph{Proceedings of the national
  academy of sciences}, vol.~99, no.~10, pp. 6562--6566, 2002.

\bibitem{deb2014integrated}
K.~Deb, S.~Bandaru, D.~Greiner, A.~Gaspar-Cunha, and C.~C. Tutum, ``An
  integrated approach to automated innovization for discovering useful design
  principles: Case studies from engineering,'' \emph{Applied Soft Computing},
  vol.~15, pp. 42--56, 2014.

\bibitem{deb2006innovization}
K.~Deb and A.~Srinivasan, ``Innovization: Innovating design principles through
  optimization,'' in \emph{Proceedings of the 8th annual conference on Genetic
  and evolutionary computation}, 2006, pp. 1629--1636.

\bibitem{yang2013stability}
P.~Yang, B.~B. Zhou, J.~Y.-H. Yang, A.~Y. Zomaya, and M.~Elloumi, ``Stability
  of feature selection algorithms and ensemble feature selection methods in
  bioinformatics,'' \emph{Biological Knowledge Discovery Handbook:
  Preprocessing, Mining and Postprocessing of Biological Data. Hoboken, New
  Jersey: John Wiley and Sons}, pp. 333--52, 2013.

\bibitem{kalra2020pan}
S.~Kalra, H.~Tizhoosh, S.~Shah, C.~Choi, S.~Damaskinos, A.~Safarpoor,
  S.~Shafiei, M.~Babaie, P.~Diamandis, C.~J. Campbell \emph{et~al.},
  ``Pan-cancer diagnostic consensus through searching archival histopathology
  images using artificial intelligence,'' \emph{NPJ digital medicine}, vol.~3,
  no.~1, pp. 1--15, 2020.

\bibitem{hegde_2019}
N.~Hegde, J.~D. Hipp, Y.~Liu, M.~Emmert-Buck, E.~Reif, D.~Smilkov, M.~Terry,
  C.~J. Cai, M.~B. Amin, C.~H. Mermel, and et~al., ``Similar image search for
  histopathology: {SMILY},'' \emph{npj Digital Medicine}, vol.~2, no.~1, 2019.

\bibitem{luigi}
D.~Komura, K.~Fukuta, K.~Tominaga, A.~Kawabe, H.~Koda, R.~Suzuki, H.~Konishi,
  T.~Umezaki, T.~Harada, and S.~Ishikawa, ``Luigi: Large-scale
  histopathological image retrieval system using deep texture
  representations,'' \emph{bioRxiv}, 2018.

\bibitem{tcga_2015}
K.~Tomczak, P.~Czerwińska, and M.~Wiznerowicz, ``Review the cancer genome
  atlas ({TCGA}): an immeasurable source of knowledge,'' \emph{Współczesna
  Onkologia}, vol.~1A, pp. 68--77, 2015.

\bibitem{bar2018chest}
Y.~Bar, I.~Diamant, L.~Wolf, S.~Lieberman, E.~Konen, and H.~Greenspan, ``Chest
  pathology identification using deep feature selection with non-medical
  training,'' \emph{Computer Methods in Biomechanics and Biomedical
  Engineering: Imaging \& Visualization}, vol.~6, no.~3, pp. 259--263, 2018.

\bibitem{tougaccar2020deep}
M.~To{\u{g}}a{\c{c}}ar, B.~Ergen, Z.~C{\"o}mert, and F.~{\"O}zyurt, ``A deep
  feature learning model for pneumonia detection applying a combination of mrmr
  feature selection and machine learning models,'' \emph{Irbm}, vol.~41, no.~4,
  pp. 212--222, 2020.

\bibitem{ozyurt2019efficient}
F.~{\"O}zyurt, ``Efficient deep feature selection for remote sensing image
  recognition with fused deep learning architectures,'' \emph{The Journal of
  Supercomputing}, pp. 1--19, 2019.

\bibitem{mirzaei2020deep}
A.~Mirzaei, V.~Pourahmadi, M.~Soltani, and H.~Sheikhzadeh, ``Deep feature
  selection using a teacher-student network,'' \emph{Neurocomputing}, vol. 383,
  pp. 396--408, 2020.

\bibitem{chen2020feature}
Z.~Chen, M.~Pang, Z.~Zhao, S.~Li, R.~Miao, Y.~Zhang, X.~Feng, X.~Feng,
  Y.~Zhang, M.~Duan \emph{et~al.}, ``Feature selection may improve deep neural
  networks for the bioinformatics problems,'' \emph{Bioinformatics}, vol.~36,
  no.~5, pp. 1542--1552, 2020.

\bibitem{khaire2019stability}
U.~M. Khaire and R.~Dhanalakshmi, ``Stability of feature selection algorithm: A
  review,'' \emph{Journal of King Saud University-Computer and Information
  Sciences}, 2019.

\bibitem{bidgoli2020collective}
A.~A. Bidgoli and S.~Rahnamayan, ``A collective intelligence strategy for
  enhancing population-based optimization algorithms,'' in \emph{2020 IEEE
  Congress on Evolutionary Computation (CEC)}.\hskip 1em plus 0.5em minus
  0.4em\relax IEEE, 2020, pp. 1--9.

\bibitem{chiew2019new}
K.~L. Chiew, C.~L. Tan, K.~Wong, K.~S. Yong, and W.~K. Tiong, ``A new hybrid
  ensemble feature selection framework for machine learning-based phishing
  detection system,'' \emph{Information Sciences}, vol. 484, pp. 153--166,
  2019.

\bibitem{rahnamayan2009center}
S.~Rahnamayan and G.~G. Wang, ``Center-based sampling for population-based
  algorithms,'' in \emph{2009 IEEE Congress on Evolutionary Computation}.\hskip
  1em plus 0.5em minus 0.4em\relax IEEE, 2009, pp. 933--938.

\bibitem{hiba2019cgde3}
H.~Hiba, A.~A. Bidgoli, A.~Ibrahim, and S.~Rahnamayan, ``{CGDE3}: An efficient
  center-based algorithm for solving large-scale multi-objective optimization
  problems,'' in \emph{2019 IEEE Congress on Evolutionary Computation
  (CEC)}.\hskip 1em plus 0.5em minus 0.4em\relax IEEE, 2019, pp. 350--358.

\bibitem{debie2019implications}
E.~Debie and K.~Shafi, ``Implications of the curse of dimensionality for
  supervised learning classifier systems: theoretical and empirical analyses,''
  \emph{Pattern Analysis and Applications}, vol.~22, no.~2, pp. 519--536, 2019.

\bibitem{bidgoli2021reference}
A.~A. Bidgoli, H.~Ebrahimpour-Komleh, and S.~Rahnamayan,
  ``Reference-point-based multi-objective optimization algorithm with
  opposition-based voting scheme for multi-label feature selection,''
  \emph{Information Sciences}, vol. 547, pp. 1--17, 2021.

\bibitem{deb2013evolutionary}
K.~Deb and H.~Jain, ``An evolutionary many-objective optimization algorithm
  using reference-point-based nondominated sorting approach, part i: solving
  problems with box constraints,'' \emph{IEEE transactions on evolutionary
  computation}, vol.~18, no.~4, pp. 577--601, 2013.

\bibitem{cooper2018pancancer}
L.~A. Cooper, E.~G. Demicco, J.~H. Saltz, R.~T. Powell, A.~Rao, and A.~J.
  Lazar, ``Pancancer insights from the cancer genome atlas: the pathologist's
  perspective,'' \emph{The Journal of pathology}, vol. 244, no.~5, pp.
  512--524, 2018.

\bibitem{pandey2017comparative}
A.~Pandey and A.~Jain, ``Comparative analysis of {KNN} algorithm using various
  normalization techniques,'' \emph{International Journal of Computer Network
  and Information Security}, vol.~11, no.~11, p.~36, 2017.

\bibitem{islam2018sample}
M.~R. Islam, ``Sample size and its role in central limit theorem (clt),''
  \emph{Computational and Applied Mathematics Journal}, vol.~4, no.~1, pp.
  1--7, 2018.

\bibitem{johanson2010initial}
G.~A. Johanson and G.~P. Brooks, ``Initial scale development: sample size for
  pilot studies,'' \emph{Educational and psychological measurement}, vol.~70,
  no.~3, pp. 394--400, 2010.

\bibitem{demvsar2006statistical}
J.~Dem{\v{s}}ar, ``Statistical comparisons of classifiers over multiple data
  sets,'' \emph{The Journal of Machine Learning Research}, vol.~7, pp. 1--30,
  2006.

\bibitem{scikit-learn}
F.~Pedregosa, G.~Varoquaux, A.~Gramfort, V.~Michel, B.~Thirion, O.~Grisel,
  M.~Blondel, P.~Prettenhofer, R.~Weiss, V.~Dubourg, J.~Vanderplas, A.~Passos,
  D.~Cournapeau, M.~Brucher, M.~Perrot, and E.~Duchesnay, ``Scikit-learn:
  Machine learning in {P}ython,'' \emph{Journal of Machine Learning Research},
  vol.~12, pp. 2825--2830, 2011.

\end{thebibliography}

\end{document}